\documentclass[lettersize,journal]{IEEEtran}
\usepackage{amsmath,amsfonts}
\usepackage{algorithmic}
\usepackage{algorithm}
\usepackage{array}
\usepackage[caption=false,font=normalsize,labelfont=sf,textfont=sf]{subfig}
\usepackage{textcomp}
\usepackage{stfloats}
\usepackage{url}
\usepackage{verbatim}
\usepackage{cite}
\usepackage{amssymb}
\usepackage{lipsum}

\usepackage[pdftex]{graphicx}
\usepackage[acronym, shortcuts]{glossaries}
\newacronym{ml}{ML}{Machine Learning}
\newacronym{ai}{AI}{Artificial Intelligence}
\newacronym{dl}{DL}{Deep Learning}
\newacronym{xai}{XAI}{eXplainable AI}

\newacronym{eo}{EO}{Earth Observation}
\newacronym{rs}{RS}{Remote Sensing}
\newacronym{scl}{SCL}{scene classification layer}
\newacronym{ads}{ADS}{additional data sources}
\newacronym{rgb}{RGB}{Red-Green-Blue}
\newacronym{dem}{DEM}{digital elevation map}
\newacronym{srtm}{SRTM}{Shuttle Radar Topography Mission}
\newacronym{aster}{ASTER}{Advanced Spaceborne Thermal Emission and Reflection Radiometer}
\newacronym{alos}{ALOS}{Advanced Land Observing Satellite}
\newacronym{ecmwf}{ECMWF}{European Center for Medium-Range Weather Forecasts}
\newacronym{sar}{SAR}{synthetic aperture radar}
\newacronym{lidar}{LiDAR}{Light Detection and Ranging}
\newacronym{fapar}{FAPAR}{Fraction of Absorbed Photosynthetically Active Radiation}
\newacronym{twi}{twi}{Topographic Wetness Index}
\newacronym{s2}{S2}{Sentinel-2}
\newacronym{sits}{SITS}{satellite image time series}
\newacronym{gsd}{GSD}{ground sampling distance}

\newacronym{ndvi}{NDVI}{normalized difference vegetation index}
\newacronym{evi}{EVI}{enhanced vegetation index}
\newacronym{vi}{VI}{Vegetation Index}
\newacronym{tsne}{t-SNE}{t-distributed Stochastic Neighbor Embedding}

\newacronym{rf}{RF}{random forest}
\newacronym{gpr}{GPR}{gaussian process regression}
\newacronym{svm}{SVM}{support vector machine}
\newacronym{mlp}{MLP}{multilayer perceptron}
\newacronym{dnn}{DNN}{deep neural network}
\newacronym{rnn}{RNN}{recurrent neural network}
\newacronym{cnn}{CNN}{convolutional neural network}
\newacronym{dfnn}{DFNN}{deep forward neural network}
\newacronym{lstm}{LSTM}{Long short-term memory neural network}
\newacronym{gbdt}{GBDT}{gradient-boosted decision tree}
\newacronym{pca}{PCA}{Principal Component Analysis}

\newacronym{mae}{MAE}{mean absolute error}
\newacronym{rmse}{RMSE}{root mean square error}
\newacronym{r2}{$\text{R}^2$}{coefficient of determination}
\newacronym{pp}{p.p}{percentage points}

\newacronym{cam}{CAM}{Class Activation Mapping}
\newacronym{gradcam}{Grad-CAM}{Gradient-weighted Class Activation Mapping}
\newacronym{shap}{SHAP}{SHapley Additive exPlanations}
\newacronym{svs}{SVS}{Shapley value sampling}
\newacronym{lime}{LIME}{Local Interpretable Model-agnostic Explanation}
\newacronym{ig}{IG}{Integrated Gradients}
\newacronym{i*g}{I*G}{Input $\times$ Gradients}

\setacronymstyle{long-short}
\usepackage[usenames,dvipsnames]{color}
\usepackage[hidelinks]{hyperref}
    \hypersetup{
        colorlinks = true,
        citecolor = {cyan},
        linkcolor =.,
        menucolor=.,
    }
\usepackage[dvipsnames,table,xcdraw]{xcolor}
\usepackage{multirow}
\usepackage{changepage}
\usepackage{soul}

\newcommand{\fct}[1]{{\mbox{\usefont{OT1}{pzc}{m}{it}{#1}}}}

\begin{document}

\title{Explainability of Sub-Field Level Crop Yield Prediction using Remote Sensing}

\author{
  Hiba Najjar,
  Miro Miranda,
  Marlon Nuske,
  Ribana Roscher,
  Andreas Dengel

\thanks{Hiba Najjar and Andreas Dengel are with the RPTU Kaiserslautern-Landau and the German Research Center for Artificial Intelligence (DFKI), Kaiserslautern, Germany (emails: najjar@rptu.de, andreas.dengel@dfki.de)}%
\thanks{Miro Miranda and Marlon Nuske are with the German Research Center for Artificial Intelligence (DFKI), Kaiserslautern, Germany (emails: miro.miranda\_lorenz@dfki.de, marlon.nuske@dfki.de)}%
\thanks{Ribana Roscher is with Forschungszentrum Jülich GmbH, Jülich, Germany and the University of Bonn, Bonn, Germany (email: ribana.roscher@uni-bonn.de).}

}

\maketitle

\begin{abstract}
    Crop yield forecasting plays a significant role in addressing growing concerns about food security and guiding decision-making for policymakers and farmers.
    When deep learning is employed, understanding the learning and decision-making processes of the models, as well as their interaction with the input data, is crucial for establishing trust in the models and gaining insight into their reliability. 
    In this study, we focus on the task of crop yield prediction, specifically for soybean, wheat, and rapeseed crops in Argentina, Uruguay, and Germany. Our goal is to develop and explain predictive models for these crops, using a large dataset of satellite images, additional data modalities, and crop yield maps.
    We employ a long short-term memory network and investigate the impact of using different temporal samplings of the satellite data and the benefit of adding more relevant modalities. 
    For model explainability, we utilize feature attribution methods to quantify input feature contributions, identify critical growth stages, analyze yield variability at the field level, and explain less accurate predictions.
    
    The modeling results show an improvement when adding more modalities or using all available instances of satellite data. The explainability results reveal distinct feature importance patterns for each crop and region. We further found that the most influential growth stages on the prediction are dependent on the temporal sampling of the input data. We demonstrated how these critical growth stages, which hold significant agronomic value, closely align with the existing literature in agronomy and crop development biology.

\end{abstract}


\section{Introduction} \label{sec:intro}
    
    \IEEEPARstart{C}{urrent} worldwide changes in the climate and political conflicts have raised concerns about food security \cite{el2022food,kemmerling2022logics}. Agriculture is the foundational block in food production, and advancements in this field are of high priority. In particular, crop yield modeling and prediction provide farmers and decision makers with valuable information to effectively manage and up-scale crop production \cite{bhanumathi2019crop,bondre2019prediction}.
    
    Data-driven techniques are being used to improve crop yield modeling, and remote sensing data is a valuable source of data for yield prediction \cite{ali2022crop}. In particular, time series of optical satellite images can trace the status of crop fields and capture the plants' growth from seeding to harvesting \cite{xie2022combining, jeong2022predicting, you2017deep, sakamoto2014near}. In this context, \gls{s2} mission emerges as a promising data source.  It consists of a constellation of two polar-orbiting satellites, with a revisit time of 5 days at the equator and 2-3 days at mid-latitudes. This temporal resolution is advantageous for crop yield prediction.
    In addition, the spatial resolution is important. The \gls{s2} resolution of 10m-30m allows for the analysis of individual fields, even at smaller sizes. It also enables the capturing of yield variability in large fields for crop production. 
    The information in the 13 \gls{s2} multi-spectral satellite images further enables the calculation of vegetation indices, revealing deeper information about the crop fields than just visible bands \cite{jin2020deep}.

    \begin{figure}[t!]
        \centering
        \includegraphics[width=0.90\columnwidth]{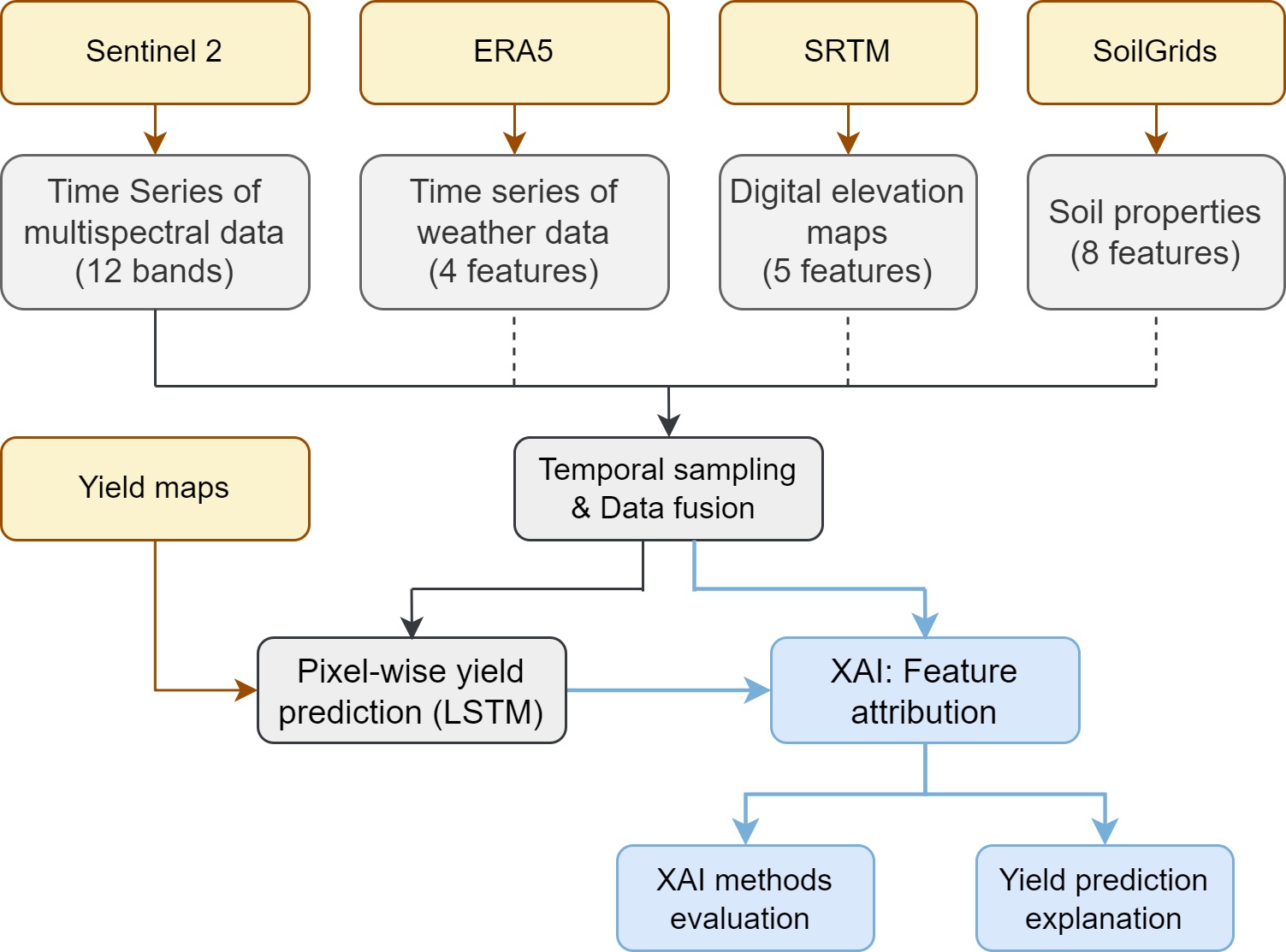}
        \caption{General workflow of the study, describing the yield prediction modeling pipeline on the top, and the explainability analysis on the bottom.}
        \label{fig:paper_overview}
    \end{figure}
    
    Leveraging the large amounts of available satellite data, \gls{ml} and \gls{dl} models have seen extensive application, as highlighted in reviews such as those conducted by Leukel et al. \cite{leukelMachineLearning2023} and Muruganantham et al. \cite{murugananthamSystematicLiterature2022}. Nevertheless, the properties of multispectral temporal images presents significant modeling challenges, often requiring the deployment of complex models. This complexity, while beneficial in many respects, does raise concerns about the transparency and interpretability of \gls{dl} models. Addressing this issue, \gls{xai} has emerged as a means to facilitate a deeper understanding of the inner workings of \gls{ml} models that often operate as black boxes.
    For example, \gls{xai} methods can support us in gaining novel insights. While agronomic research studies have improved our understanding as humans of crop yield drivers and growth factors, this knowledge might not necessarily align with what a \gls{ml} model would learn from satellite data \cite{neuhaus2023spurious}. In particular, \glspl{dnn} might learn novel, insightful patterns when relating the satellite image input to the yield output, which might have been unknown to the field of agronomy so far. On the other side, these models might learn spurious correlations, hindering their capacity to generalize once they are deployed in production pipelines. \gls{xai} methods can support here to identify errors in the learned model \cite{lime_ribeiro2016should}.
    
    Another important advantage of explainable yield prediction models lies in their reliability when applied to real-world scenarios. For instance, these models can prove invaluable to insurance companies in the aftermath of a loss, serving as a benchmark against which actual yields can be measured. Moreover, they can help in decision-making processes by assessing the potential risk associated with insuring a particular agricultural operation.
    Farmers, too, can gain significantly from the implementation of such yield prediction models. These tools can aid farmers in managing risks associated with crop production, and  optimizing the use of resources such as water, fertilizers, and pesticides. The transparency offered through the explanation of the model's functioning adds an extra layer of credibility, resulting in an increased trust among users in the accuracy of its predictions.

    In this paper, we train and explain crop yield prediction models using a large dataset of yield maps collected from three countries for three different crop types. 
    Although we do not focus on advancing the yield modeling task,
    our primary objective is to employ feature attribution methods \cite{ig_sundararajan2017axiomatic} to interpret individual predictions and explain them through domain knowledge. Additionally, we aggregate the results at the different levels to gain deeper insights into the spectral bands and growth stages that the model uses and their impact on the yield prediction accuracy.
    We summarize the workflow of data preparation, yield modeling, and model explanation in Figure \ref{fig:paper_overview}.
    The main contributions of our work include: 
    
    \begin{itemize}
    \setlength\itemsep{0.1em}
        \item We explore the potential of a higher level of interpretability when replacing raw satellite bands with vegetation indices; 
        \item We evaluate and compare eight attribution methods to decide on a robust technique, which we use to identify the most influential features in our training data;
        \item We analyze the attribution results on all experiments and aggregate them on different levels to identify region- and crop-specific spectral and temporal features that are important for the model to predict the yield;
        \item We explain the in-field variability and explore the correlation between the feature importance distribution and the performance of the model on the field level;
        \item Finally, we conducted a thorough review of existing literature on crop growth and development to explain our attribution results.
    \end{itemize}

\section{Related Work} \label{sec:rel_work}

    In the literature, some studies using remotely sensed data addressed the interpretability concern and explained their yield prediction model.
    In the following, we discuss studies using remotely sensed data for yield prediction that address the explanation of the model, distinguishing between papers using classical \gls{ml} models and more advanced deep architectures. 
    
    Starting with the utilization of classical \gls{ml} models, 
        Martínez-Ferrer et al. \cite{martinez-ferrerCropYield2021} train a \gls{rf}, linear regression and \gls{gpr} models to predict soy, corn and wheat yield across the continental US. 
        They combine three satellite-based bio-geophysical variables with additional meteorological features and use occlusion-based sensitivity to estimate the importance of each predictive feature.
        Farmonov et al. \cite{farmonovCombiningPlanetScope2023} predict the wheat yield for a single harvesting season across four fields in Hungary. 
        They use satellite images from PlanetScope and \gls{s2}, along with vegetation indices, meteorological terrain variables. 
        A \gls{rf} model is trained to predict the yield at the pixel level, and feature importance is estimated using the variable-specific node impurity method inherent to ensemble tree-based models.
        Broms et al. \cite{bromsCombinedAnalysis2023} train 30 different tree-based models to predict winter wheat yield from 23 farms in southern Sweden, on a single harvesting season. 
        Sentinel-2 multispectral data, derived indices, soil properties, weather data, terrain variables and farming information were processed by the models. 
        A feature importance analysis is then conducted through the estimation of Shapley values \cite{shapley1953value, svs_strumbelj2010efficient}.
        Celik et al. \cite{celik2023explainable} predict cotton yield using an extension of the generalized additive models \cite{nori2019interpretml} and integrating satellite remote sensing images, climate data, and soil parameters. They utilize the intrinsic interpretability of the model to extract the interactions between the features, their importance, and their interpretation.

    Examining studies that employ deep networks,
        Wolanin et al. \cite{wolaninEstimatingUnderstanding2020} train a \gls{cnn} on time series comprising vegetation indices and meteorological data to predict wheat yields in the Indian Wheat Belt. 
        Their dataset counts 143 districts and is spanned over a 13-year period. After training the model, regression activation maps are implemented to visually explain the model and identify significant time steps.
        At the county level, Huber et al. \cite{huberExtremeGradient2022} estimate soybean yield using MODIS bands and weather data. 
        They compare the performance of XGBoost, \gls{cnn} and \gls{cnn}-\gls{lstm} models for the modeling phase. 
        After model training, Shapley values were computed to estimate and analyze feature importances.
        Paudel et al. \cite{paudel2023interpretability} compares early and intermediate fusion techniques to combine temporal and static data for crop yield prediction. 
        Their predictive variables include time series of remotely sensed weather data, crop model indicators \cite{de201925} and the \gls{fapar} \cite{FAPAR}, in addition to static features, such as agro-environmental zone identifier, terrain elevation features, soil moisture and field size. 
        The authors train a gradient-boosted decision tree model, a \gls{lstm} network and a 1D-CNN. The \gls{lstm} model is then explained using Shapley values.
        Isik et al. \cite{isik2023interpretable} train a similar model and explanation method to predict cotton yield in three regions in Turkey at the county level. They use \gls{evi}, climate variable and soil properties covering the period from 2018 until 2022. 

    While most of the aforementioned studies analyse the influence of the input features on the predictions, none can claim accurate feature importance since they estimate yield at a regional (county or district) level. 
    This limitation arises from the difficulty of obtaining yield map data, which is not always readily available, while collecting information about the total harvest for a field or region is more accessible.
    However, Farmonov et al. \cite{farmonovCombiningPlanetScope2023} stand out as an exception. They use combine harvester yield maps as ground truth to predict yield at the sub-field level. 
    Nevertheless, their training data is limited in size, consisting of pixels from only two wheat fields, both from the same experimental farm. Furthermore, they used a simple statistical model, a \gls{rf}, which is oblivious to the temporal dimension and may, therefore, fail to capture complex temporal patterns in the data as effectively as \glspl{dnn} can.

    In contrast, our study utilizes a large yield map dataset comprising 1061 fields across three different crops (soybean, wheat, and rapeseed) from three countries (Uruguay, Argentina, and Germany). 
    We analyze time series data from multispectral satellite images, supplemented by additional data sources including soil, terrain, and weather. Our approach involves predicting yields at the subfield level by processing the data pixel-wise. We employ an \gls{lstm} architecture, which has already demonstrated success in analyzing temporal data \cite{mateo-sanchisInterpretableLongShort2023,paudel2023interpretability,russwurm2020self,hochreiter1997long}.
    We then employ \gls{xai} techniques and combine it with domain knowledge to extract explanations for the predictions of these models. We aggregate local explanations for feature importance to generate a global explanation at different levels. 
    This comprehensive approach allows us to analyze both temporal and spectral importance, providing valuable insights into the inner workings of the trained networks and enabling comparisons of the model's behavior across different crops and regions.
    We further validate the results against known yield factors from agronomic studies in the literature. 
    To the best of our knowledge, our work represents the first comprehensive explanation of yield prediction using a large remote sensing and yield maps dataset at the sub-field level. 
\section{Datasets and Methods}\label{sec:methodology}

    \subsection{Study sites}
\begin{figure}[h]
            \centering
            \includegraphics[width=0.80\columnwidth]{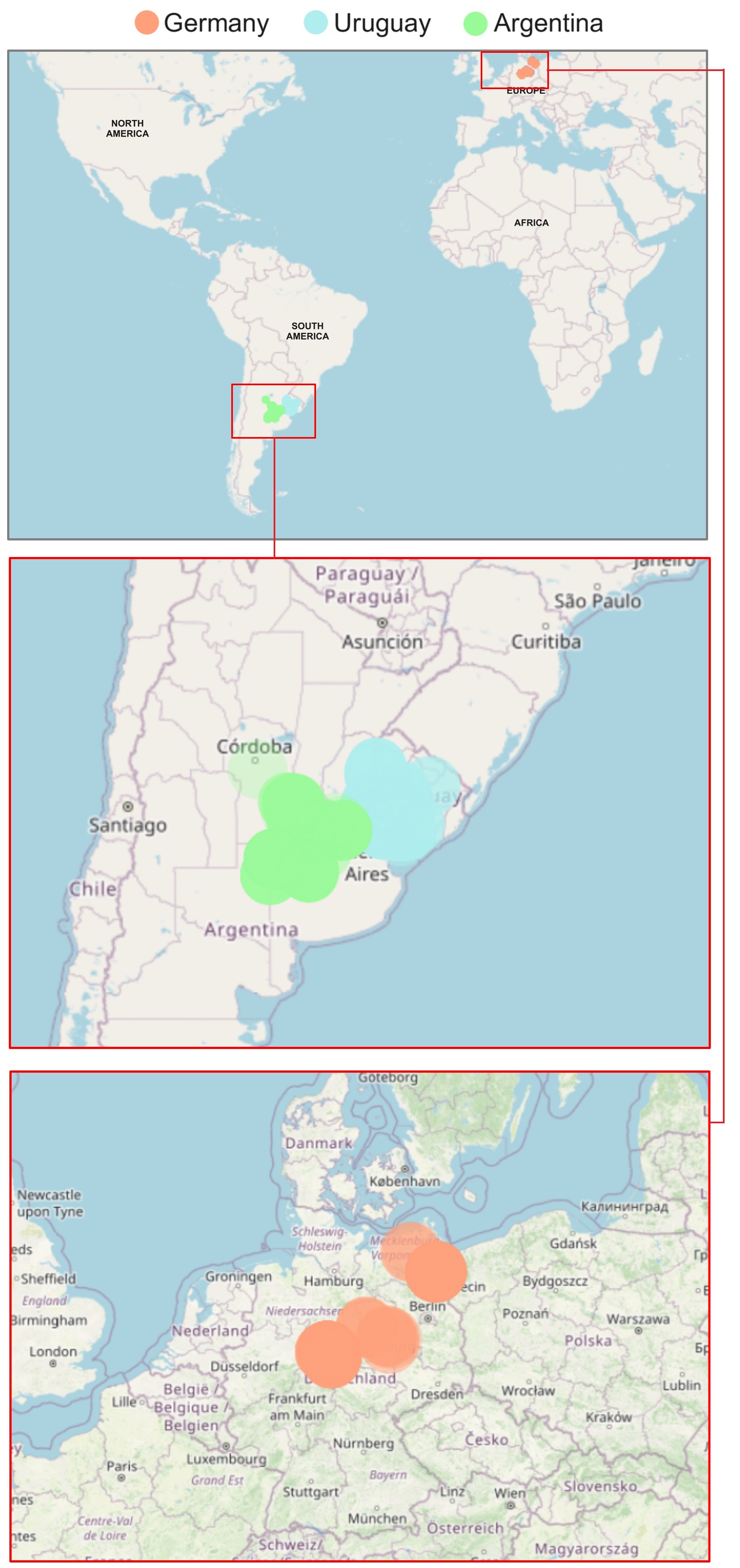}
            \caption{Geolocations of crop fields from South America and Europe: Argentina (soybean), Uruguay (soybean) and Germany (wheat and rapeseed). 
            Map data copyrighted OpenStreetMap contributors.}
            \label{fig:ss_maps}
        \end{figure}

        We collected agricultural data from different regions in South America and Europe, as depicted in Figure \ref{fig:ss_maps}. Specifically, we gathered soybean crop data from two countries: Argentina and Uruguay, involving 10 and 8 different farms, respectively. In Northern Germany, we used data from wheat and rapeseed crop fields, all originating from the same 6 farms.
        In Argentina, all soybean farms used in our study are situated in the northern regions, close to Uruguay, while in Germany, rapeseed and wheat crops are cultivated in the same farms, located in the northwest and close to the center of the country.

        To gain a comprehensive understanding of the climate conditions and topography in these regions, we rely on the Global Agro-Ecological Zones (GAEZ) model documentation, specifically for insights into land, water \footnote{Available from \href{https://gaez-services.fao.org/apps/theme-1/}{https://gaez-services.fao.org/apps/theme-1/}, accessed 20 March 2024.} and agro-climatic \footnote{Available from \href{https://gaez-services.fao.org/apps/theme-2/}{https://gaez-services.fao.org/apps/theme-2/}, accessed 20 March 2024.} ressources \cite{fischer2021global}.
        All our study regions share a sub-tropic climate with moderately cool temperatures. Furthermore, the air is humid in North Argentina and Uruguay, sub-humid without significant soil or terrain constraints in German fields near the country's center, and semi-arid with notable soil and terrain limitations in northern Germany.
        The total number of rain days (days with daily precipitation exceeding 1 mm) varies across these regions. In North Argentina and Uruguay, it ranges from 100 to 130 rain days, with annual precipitation averaging around 1900mm and 1000-1900mm, respectively. In Northern Germany, the number of rain days varies from 160 to 220, with annual precipitation ranging between 500 and 700mm.
        The length of the growing season is approximately 200 days in Germany and extends to 300 days in North Argentina and Uruguay.
        Note that the annual precipitation values and growing period lengths are averaged over the period 1981-2010.
        As for the topography of the study sites, the median altitude exhibits significant variations. In German fields, altitudes range from 40m in the northwest to around 250m near the country's center. In Uruguay, altitudes span from 30m to 200m, while in Argentina, the altitude variation is relatively smaller, ranging from 30m to 120m.

    \subsection{Data and Yield Modeling}
        
        \begin{table*}[t]
\caption{Overview of the spatial and temporal coverage and the size of the yield map datasets.}
\centering
\begin{tabular}{ccccccc}

\hline 
Dataset & Region & Crop Type & Years &  \# Farms & \# Fields & \# Pixels \\
\hline 
ARG-S & Argentina & soybean & $2017-2022$ & 8 & 190 & 1.45M \\
URG-S & Uruguay & soybean & $2018-2022$ & 10 & 572 & 1.79M\\
GER-R & Germany & rapeseed & $2016-2022$ & 6 &  111 & 0.30M\\
GER-W & Germany & wheat & $2016-2022$ & 6 & 188 & 0.31M\\
\hline
\end{tabular}
\label{table:datasets}
\end{table*}

        In our study, we apply \gls{xai} to the \gls{ml}-based crop yield prediction task on the subfield level. The target dataset consists of rasterized yield maps with a resolution of 10 meters, which we divided into four datasets, as outlined in Table \ref{table:datasets}. It is important to note that a field can be counted more than once if its yield data was collected for multiple years. 
        Uruguay provides the largest amount of soybean fields for our study, primarily due to the number of farms involved, followed by Argentina and Germany. The field area varies from one country to another, which explains the disproportionality between the numbers of fields and pixels.
        To clean the yield data, we implemented a yield thresholding at 15t/h on the yield maps, as recommended by experts in agriculture. We subsequently applied an outlier filter using the three-sigma rule, removing data points falling beyond the range defined by three standard deviations from the mean. Further cleaning included an inter-quartile-range filter, as detailed in \cite{sanchezInfluenceData2023}.
        
        To predict the yield, separate models are trained for each dataset. We supply the model mainly with satellite data, and conduct additional experiments adding more modalities to the network.  More specifically, we use the entire spectral signal of Sentinel-2 Level-2A data and collect all scenes from seeding to harvesting. In terms of \gls{ads}, and following the same process as in \cite{pathakPredictingCrop2023}, we collect information about soil condition, \glspl{dem}, and weather data. All maps were rasterized to match the yield resolution (10m). 
        In Appendix \ref{app:data}, a summary of the predictive features used is shown in Table \ref{table:features}, along
        additional information about the processing of input and output data.

    \subsection{Interpreting Yield Modeling through Feature Attribution}
        
        Feature attribution methods are commonly used interpretation techniques that, when combined with domain knowledge, provide explanations for individual predictions. An attribution method assigns sensitivity or contribution scores to the input features, with the absolute value of these scores quantifying their importance on the model's prediction. In our paper, we will evaluate and employ such techniques to explain the yield prediction models.
        
        Given the $i^{\text{th}}$ pixel sample $X^i \in \mathbb{R}^{B \times T}$, where $B$ and $T$ are the numbers of bands and time steps, respectively, let $a^i_{b,t} \in \mathbb{R}$ represent the attribution result for the feature $x^i_{b,t}$. 
        The relative value of $a^i_{b,t}$ indicates the extent to which this feature contributed to increasing or decreasing the final predicted value $\hat{y}^i \in \mathbb{R}$.
        
        \paragraph{Preprocessing of raw attributions} 
        In this paper, all attribution results systematically go through two types of transformations before any further analysis. 
        The first transformation, $\hat{a}^i_{b,t}$, is used to create attribution maps, where we scale the attributions across all the features of each pixel so that the total absolute attribution is equal to $1$. 
        The second transformation, $\bar{a}^i_{b,t}$, which we apply in all the remaining analyses, uses the absolute values of the first operation. It reflects the absolute importance to the model's prediction, which is exactly what we are assessing in this work:

        \begin{equation}\label{eq:attr_scaling}
            \hat{a}^i_{b,t} = \frac{ a^i_{b,t} }{ \sum_b \sum_t \left| a^i_{b,t} \right| }, \quad
            \bar{a}^i_{b,t} = \left| \hat{a}^i_{b,t} \right|,  
        \end{equation}

        where $t \in \{ 0,1, \dots, T \}$ and $b \in \{ 0,1, \dots, B \}$. In addition, we can evaluate the spectral importance $\fct{SI}(X^i)_b$ and temporal importance $\fct{TI}(X^i)_t$ of a given band $b$ and time step $t$, respectively, by summing up the transformed attributions along the temporal or spectral dimension:

        $$
        \fct{SI}(X^i)_b = \sum_{t} \bar{a}^i_{b,t}, \forall b \in \{ 0,1, \dots, B \};
        $$
        $$
        \fct{TI}(X^i)_t = \sum_{b} \bar{a}^i_{b,t}, \forall t \in \{ 0,1, \dots, T \} . 
        $$

        Furthermore, the importance of a specific band or time step can be evaluated on the field level by averaging its importance over all the pixels from that field, and similarly at the dataset level. \\[0.3cm]
        
        \paragraph{Attribution methods comparison} 
        In order to achieve insightful explanations, we compare eight commonly used attribution methods and select the most suitable one for our further analysis. As a follow-up work of our initial results in \cite{najjarFeatureAttribution2023}, we compare the results of the following methods: Saliency \cite{saliency_simonyan2014deep}, \gls{ig} \cite{ig_sundararajan2017axiomatic}, \gls{i*g} \cite{shrikumar2016not}, GradientShap \cite{shap_lundberg2017unified}, Occlusion \cite{occ_zeiler2014visualizing}, \gls{svs} \cite{svs_strumbelj2010efficient}, \gls{lime} \cite{lime_ribeiro2016should} and KernelShap \cite{shap_lundberg2017unified}. 
        The spectral and temporal importances of the interpreted samples are compared qualitatively.  Moreover, additional attribution scores, which are the sensitivity and infidelity metrics \cite{yeh2019fidelity}, are computed for a quantitative evaluation. 
        Each score assigns a single value for each data sample to assess the quality of the computed attribution.
        They quantify the instability of the attribution of an input $X$ by perturbing this input into $X^\prime$, computing the attribution of $X^\prime$, and comparing both attributions. 
            The sensitivity score evaluates the degree to which the explanation is affected when the input is slightly perturbed, 
            while the infidelity score computes the expected difference between (i) the dot product of the input perturbation to the explanation vector $\mathbf{a}$, and (ii) the outcome difference between $X$ and $X^\prime$.\\[0.3cm]
        
        \paragraph{Attribution baselines} 
        Most of the attribution methods require a user-defined baseline vector, which can serve different purposes: it is used by \gls{lime}, Occlusion, \gls{svs} and KernelShap to replace occluded features at computation time, and it serves as a starting point from which the integral in \gls{ig} and the gradients expectation in GradientShap are computed.
        In this work, we introduce and compare two distinct baselines for the attribution methods, which we term the \textit{mean baseline} and the \textit{padded baseline}.
        In contrast to a similar study which defines the mean baseline as the mean vector $\bar{X}$ computed across the entirety of each dataset \cite{najjarFeatureAttribution2023}, in this study, we further adapt this vector to each input pixel, $\bar{X}^{\text{base}(i)}$, using the following rule, for each band $b$ and time step $t$:

        \begin{equation*}
            \bar{X}^{\text{base}(i)}_{b,t}=
            \begin{cases}
              -1, & \text{if}\ X^{i}_{b,t}=-1 \\
              \bar{X}, & \text{otherwise.}
            \end{cases}
        \end{equation*}
        
        This adaptation addresses a critical consideration: the varying presence of missing satellite data at some early and/or late time steps within each field, which are padded in our experiments with the value $-1$. To counteract this, we align the missingness pattern of each pixel with the corresponding segment within the mean vector. This adapted mean vector then serves as the reference when calculating the attribution values of that pixel.
        On the other hand, the padded baseline essentially entails a vector that is fully padded with the $-1$ value.
        
        A careful definition of the baseline vector holds particular significance due to the nature of the perturbation-based attribution methods, which involve occluding a feature to deduce its significance. In these methods, the occlusion process substitutes the feature's value with the corresponding value from the baseline vector. 
        In scenarios where the occluded value stems from a missing time step, but is substituted with a high value taken from the mean vector $\bar{X}$, a substantial value shift occurs. 
        This can potentially distort predictions in a misleading manner, implying a high importance of the occluded feature. 
        By using the adjusted baseline vector $\bar{X}^{\text{base}(i)}$, any absent feature maintains a consistent padding value when subjected to occlusion, resulting in its importance remaining effectively negligible.

    \subsection{Interpretable features for Interpretable models}

        As discussed in \cite{guidotti2018survey}, the process of explaining a model through feature attribution methods requires a solid comprehension of the predictive features being employed. Conventional methodologies for crop modeling rely on the crafting of features by experts, with the aim of encapsulating the extensive array of variables recognized as direct yield drivers. However, when \gls{ml} and \gls{dl} models are used, the scenario changes. Raw information, acknowledged for its potential to serve as a proxy for crop growth and health, is directly supplied to the network with minimal to no alterations. In our work, we address this limitation of explaining \gls{ml} models in two dimensions: inclusion of vegetation indices and aggregation of the temporal dimension.
        
        With the aim of augmenting the interpretability of the spectral dimension, we employ well-established vegetation indices, frequently employed in the literature for crop classification and yield prediction. These vegetation indices merge two or more spectral bands and generate an index with agronomic relevance, rendering it more easily interpretable than the raw bands. The fundamental idea entails substituting the original \gls{s2} raw bands with pre-calculated indices during the training phase. Subsequently, we will evaluate the impact of this adjustment on the model's performance.
        In our experiments, we will investigate the impact on accuracy when training the model using vegetation indices and compare it to training the model using raw inputs. If the results are similar, we will continue to use the model trained with vegetation indices for future experiments. However, if there is a significant decrease in performance, we will not further use the improvement in spectral interpretability.
        The full list of the indices used and their respective formulas are summarized in Appendix \ref{app:VIs}.
        
        To gain insights and derive explanations from crop yield models using time series, we can use different types of domain knowledge, with particular potential in the different phases that each crop goes through.
        To enhance the temporal interpretability of the input feature and derive explanations from it, we will exploit the principal growth stages of the BBCH scale for rapeseed and wheat \cite{lancashire1991uniform}, and a modified scale for soybean \cite{mcwilliams1999soybean}, as summarized in Appendix \ref{app:growth_stages}. 
        Concretely we first get the modified attributions $\bar{a}^i_{b,t}$ for each explained pixel sample $X^i$. We further define for each growth stage $t_{\text{gs}}$ a set $T_{\text{gs}}$ of the time steps it covers, then compute the total attribution as follows:
        
        $$
        \bar{a}^i_{b,t_{\text{gs}}} = \sum_{t \in T_{\text{gs}}} \bar{a}^i_{b,t} 
        $$
        
        A major benefit of this approach is that it overcomes the misalignment of the seeding and harvesting months between the different fields in each dataset, which would have hindered the interpretability of the raw attributions on a higher level. 

\section{Experimental Setup}\label{sec:setup}

        \paragraph{Time Series Modeling} 
        To process the data pixel-wise, we create a multivariate time series for each pixel of the rasterized data maps.  For the temporal sampling, we compare two approaches: \textit{raw time-series} and \textit{monthly sampling}.
            In the former approach, all the satellite data available between seeding and harvesting dates is used. Across all datasets, the growth period and available data depend on the crop type and the latitude coordinate of the field, which results in different revisit times: countries closer to the equator have a longer gap between two consecutive satellite images compared to farther countries. Thus, we fix the sequence length to 150 across all datasets and pad the initial time steps in case of shorter sequences.
            
            In the monthly sampling, the sequence length is fixed to 24 months, corresponding to two calendar years, such that the harvesting date always falls in the second year. 
            This strategy is implemented to maintain unique indices for each month across various fields and years. However, only data from seeding to harvesting are collected, while the remaining time steps are padded \cite{pathakPredictingCrop2023}.
            In general, using a uniform sequence length further facilitates the processing of the extensive dataset in mini-batches during model training and validation.
            When training with additional modalities under the monthly sampling scenario, static data (i.e. soil and \gls{dem}) are duplicated at each time step, while weather data is monthly aggregated by computing the sum of each weather feature, before being stacked to the satellite time series.

        \paragraph{Yield modeling}
            In order to capture the temporal dependencies in the data, we use \gls{lstm} networks and train a separate model for each dataset, under three different scenarios: raw time-series, monthly sampling, and monthly sampling with additional modalities.
            The training is conducted under the cross-validation setting using 10 folds, grouped by fields and stratified by farms, to ensures that pixels from the same field are always grouped in the same fold.
            To evaluate the quality of the prediction and the model performance, \gls{mae}, \gls{rmse}, and the \gls{r2} metrics are used.
            In Appendix \ref{app:model}, we describe the architecture of the \gls{lstm} models and the formulas of the metrics used.

        \paragraph{Attribution methods comparison}
            The soybean dataset from Argentina is used for the comparison of multiple feature attribution methods, and the data is preprocessed according to the monthly sampling. 
            To use explainability for scientific insights, higher model accuracies yield more informative and stable models. We specifically interpret the fold with the highest \gls{r2} score, and compute the attribution of all the pixels in the fold-specific validation fields.
            
        \paragraph{Interpreting yield prediction models}
            Since perturbation-based attribution methods involves repetitive perturbations of the input and model prediction evaluations, interpreting the entire datasets can be computationally expensive.
            However, to ensure representative explanation results, we employ a random selection process where we pre-select a maximum of five fields per farm and explain all pixels within these fields.
            Furthermore, from each dataset, we identify and interpret the model from the cross-validation fold that achieved the highest \gls{r2} scores across the three training scenarios. 
            
\section{Results}\label{sec:results}

    \subsection{Yield Prediction}\label{subsec:res_yield_pred}

        We first evaluate the performance of the yield prediction models, where one model is trained for each dataset and for each sampling and feature composition setting. The results, based on the evaluation metrics used, are summarized in Tables \ref{table:model_raw_ts_eval} and \ref{table:model_monthly_eval} for both the raw and monthly samplings, respectively.

        We conduct the evaluation at both the subfield and field levels. For the former, we compute each metric for each pixel and then average over the entire dataset. In the latter case, we first average the predicted and target yield values over the pixels within each field. Then, we compute the metrics for each field and finally, we average them over the entire dataset.

\begin{table}[h]
\footnotesize
\caption{Modeling results under the raw time series on all datasets and with two different training data: satellite bands (S2) or vegetation indices (VI). The best scores in each dataset are highlighted in bold.}
\centering 
\begin{tabular}{c|c|ccc|ccc}
\multirow{2}{*}{\textbf{Dataset}} & \multirow{2}{*}{\textbf{Bands}} & \multicolumn{3}{c|}{\textbf{Field}} & \multicolumn{3}{c}{\textbf{Subfield}} \\
 &  & \textbf{MAE} & \textbf{R2} & \multicolumn{1}{c|}{\textbf{RMSE}} & \textbf{MAE} & \textbf{R2} & \multicolumn{1}{c}{\textbf{RMSE}} \\
\hline
\multirow{2}{*}{ARG-S} 
 & VI & 0.41 & 0.72 & 0.55 & 0.67 & 0.62 & 0.90 \\
 & S2 & \textbf{0.38} & \textbf{0.76} & \textbf{0.52} & \textbf{0.66} & \textbf{0.64} & \textbf{0.88} \\
 \hline
\multirow{2}{*}{URG-S}
 & VI & \textbf{0.36} & \textbf{0.77} & \textbf{0.51} & \textbf{0.78} & \textbf{0.41} & \textbf{1.22} \\
 & S2 & \textbf{0.36} & 0.75 & 0.53 & \textbf{0.78} & \textbf{0.41} & \textbf{1.22} \\
\hline
\multirow{2}{*}{GER-R} 
 & VI & 0.71 & 0.53 & 0.92 & 1.04 & 0.31 & 1.38 \\
 & S2 & \textbf{0.61} & \textbf{0.67} & \textbf{0.77} & \textbf{0.98} & \textbf{0.38} & \textbf{1.31} \\
\hline
\multirow{2}{*}{GER-W} 
 & VI & 1.45 & 0.11 & 1.79 & 2.21 & 0.07 & 2.84 \\
 & S2 & \textbf{0.91} & \textbf{0.55} & \textbf{1.27} & \textbf{1.73} & \textbf{0.35} & \textbf{2.38} \\
\end{tabular}
\label{table:model_raw_ts_eval}
\end{table}
        
\begin{table}[h]
\footnotesize
\caption{Modeling results under the monthly temporal sampling on all datasets and with three different training data: satellite bands (S2), satellite bands and additional modalities (S2+ADS) and vegetation indices (VI). The best scores in each dataset are highlighted in bold.}
\centering 
\begin{tabular}{c|c|ccc|ccc}
\multirow{2}{*}{\textbf{Dataset}} & \multirow{2}{*}{\textbf{Bands}} & \multicolumn{3}{c|}{\textbf{Field}} & \multicolumn{3}{c}{\textbf{Subfield}} \\
 &  & \textbf{MAE} & \textbf{R2} & \textbf{RMSE} & \textbf{MAE} & \textbf{R2} & \textbf{RMSE} \\ 
 \hline
\multirow{3}{*}{ARG-S} & VI & 0,46 & 0,64 & 0,63 & 0,74 & 0,55 & 0,98 \\
 & S2 & 0,4 & 0,74 & 0,53 & 0,69 & 0,61 & 0,92 \\
 & S2+ADS & \textbf{0,38} & \textbf{0,76} & \textbf{0,51} & \textbf{0,66} & \textbf{0,63} & \textbf{0,89} \\
 \hline
\multirow{3}{*}{URG-S} & VI & 0,44 & 0,64 & 0,64 & 0,82 & 0,36 & 1,27 \\
 & S2 & 0,4 & 0,69 & 0,59 & 0,8 & 0,38 & 1,25 \\
 & S2+ADS & \textbf{0,35} & \textbf{0,76} & \textbf{0,52} & \textbf{0,78} & \textbf{0,41} & \textbf{1,23} \\
 \hline
\multirow{3}{*}{GER-R} & VI & 0,62 & 0,64 & 0,81 & 1,02 & 0,34 & 1,36 \\
 & S2 & 0,6 & 0,65 & 0,8 & 1,01 & 0,35 & 1,35 \\
 & S2+ADS & \textbf{0,55} & \textbf{0,69} & \textbf{0,75} & \textbf{0,96} & \textbf{0,41} & \textbf{1,29} \\
 \hline
\multirow{3}{*}{GER-W} & VI & 1,31 & 0,23 & 1,66 & 2,07 & 0,17 & 2,68 \\
 & S2 & \textbf{0,91} & 0,57 & 1,25 & 1,79 & 0,31 & 2,45 \\
 & S2+ADS & 0,92 & \textbf{0,59} & \textbf{1,22} & \textbf{1,74} & \textbf{0,33} & \textbf{2,41} \\
\end{tabular}
\label{table:model_monthly_eval}
\end{table}

        In all our experiments, we consistently observe that the results at the field level outperform those at the subfield level. This disparity arises because subfield metrics measure the model's ability to capture the in-field variability of yield values. In contrast, on the field level, we notice that averaging the individual predictions can more easily and accurately estimate the average yield per field.
        
        When comparing the \gls{r2} results across the different datasets, we note that the top scores at the field level are generally comparable across the datasets, except for GER-W, which consistently ranks last by a considerable margin. 
        At the subfield level, the results achieved in ARG-S clearly outperform those in other datasets, regardless of the temporal samplings used, while GER-W is still not reaching comparable scores. 
        These disparities may originate from the distinct agronomic characteristics of each crop. Therefore, it is possible that the input data in our experiments might be inherently insufficient to capture these unique characteristics and establish a strong connection with the target yield, resulting in less accurate predictions.
        Furthermore, satellite data availability can also vary from one country to another, influenced by factors such as geolocation and cloud coverage related to the climate in each region \cite{sudmanns2020assessing}. 
        The quality of the yield data collected from different countries may also further impact the overall results.

        However, when replacing the full spectrum of satellite data with vegetation indices, we observe different responses. 
            In the monthly sampling, this change had a minor negative impact in URG-S and GER-R, but in contrast, it led to a substantial decrease in scores in ARG-S and GER-W datasets. 
            Notably, in these two datasets, the \gls{r2} score dropped by 10 and 34 \gls{pp}, respectively, at the field level.
            When applying this same operation to raw time series data, a similar decrease in scores is observed in Germany's datasets, with a 14 \gls{pp} drop in rapeseed fields and a 44 \gls{pp} drop in wheat fields at the field level. 
            However, the use of vegetation indices yielded comparable results to the full satellite spectrum in Argentina and even slightly improved results at the field level in Uruguay. 
            One reason for these results might be the need for complex spectral band interactions. Vegetation indices usually map only linear or low-complexity inter-band interactions, while neural networks are able to learn rich non-linear interactions.
            
        Analyzing the impact of enriching the satellite data with additional modalities, as shown in Table \ref{table:model_monthly_eval}, reveals an overall improvement in the results across all datasets. This results also align with the literature \cite{mena2023common}. 
        The smallest increase is observed in the Argentina dataset, while the most significant impact is seen in the rapeseed dataset in Germany.
    
    \subsection{Feature Attribution Methods}

\begin{figure}[h]
            \centering
            \includegraphics[width=1\columnwidth]{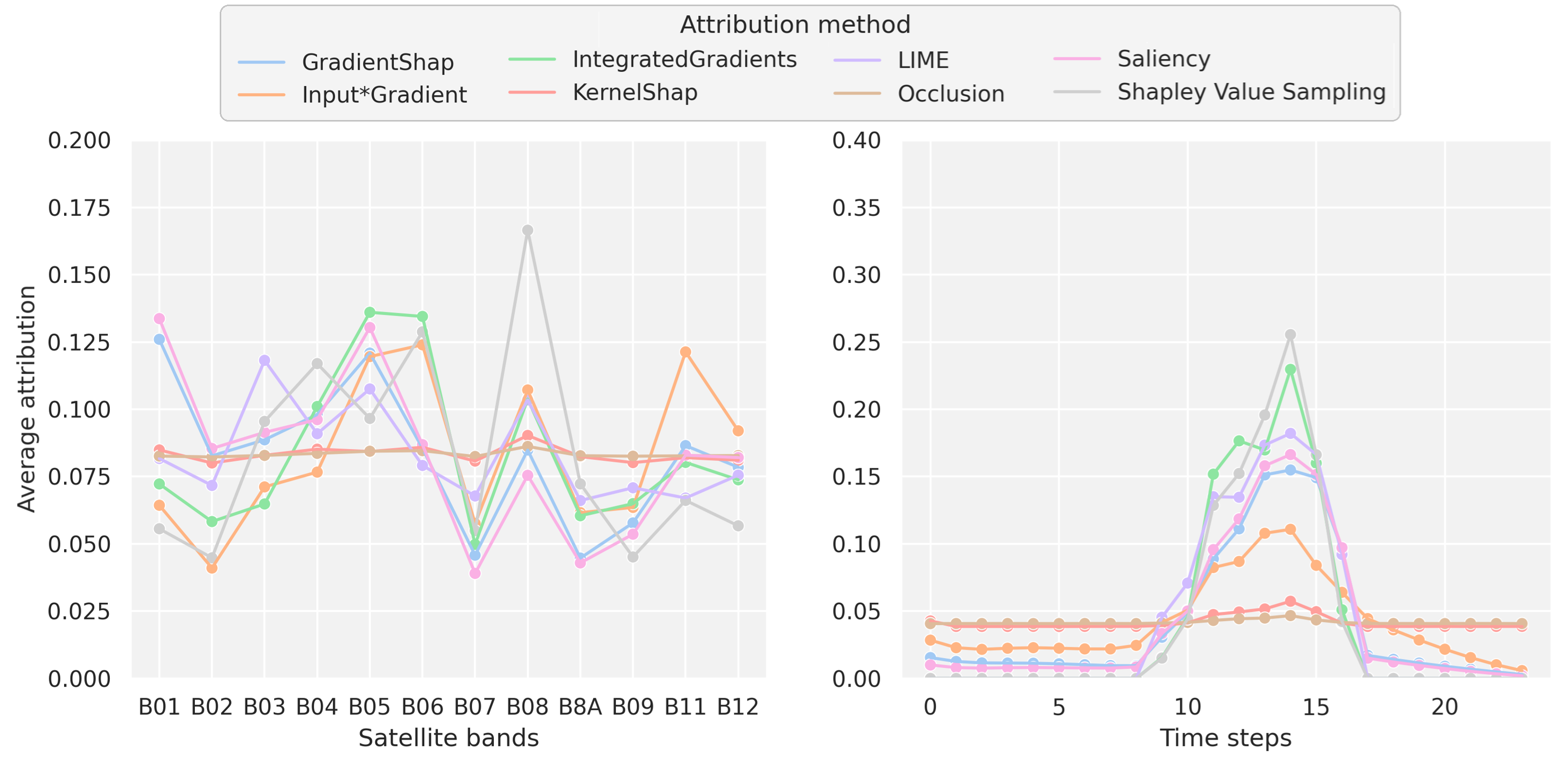}
            \caption{Feature attribution methods results using the mean baseline and ARG-S dataset. On the left is the spectral importance, and on the right is the temporal importance.}
            \label{fig:mean_baseline_spectral_temporal_importance}
        \end{figure}

        To identify a robust attribution method for the interpretation of the yield prediction models, we conducted a comparison of different techniques using the same dataset and model. Using the mean baseline vector, we compute both spectral and temporal importances, as illustrated in Figure \ref{fig:mean_baseline_spectral_temporal_importance}.
        We observe a consistent pattern in the temporal attributions. Specifically, time steps before seeding and after harvesting months (approximately from time step 9 to 16) exhibited low importances, while the early growing months had a more significant influence, and the late stages were the most important.
        However, looking at the spectral importance, it is hard to decipher an agreement pattern across the feature attribution methods.
        This highlights the need for further evaluation tools to assess these methods better.

\begin{figure}[h]
            \centering
            \includegraphics[width=0.70\columnwidth]{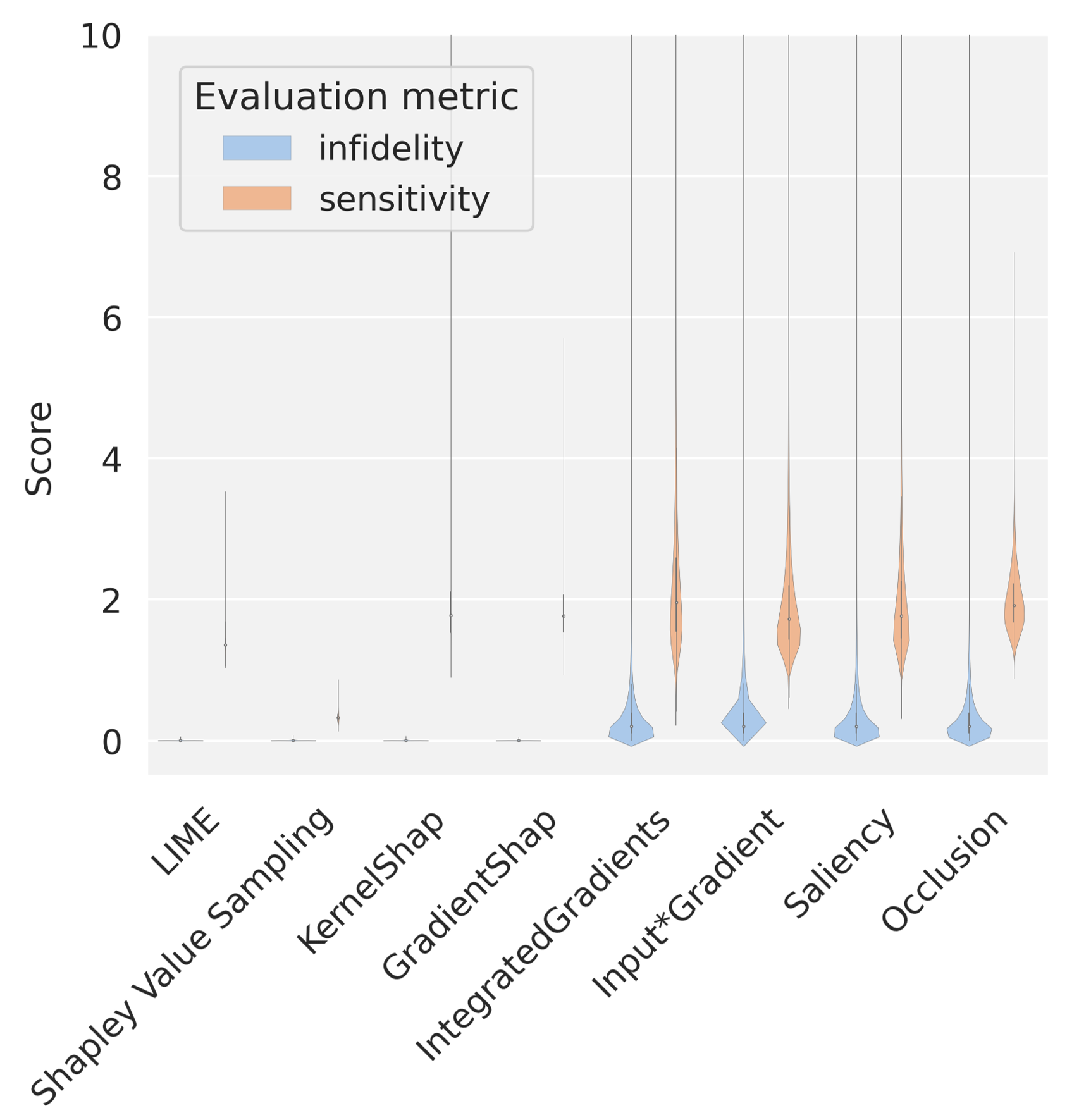}
            \caption{
                Quantitative evaluation of the feature attribution methods using the mean baseline on ARG-S dataset: Infidelity and sensitivity scores.
                }
            \label{fig:mean_baseline_eval_scores_b}
        \end{figure}

        In Figure \ref{fig:mean_baseline_eval_scores_b}, we assess the stability of the methods and their robustness against both minor and significant input modifications.
        This is quantified using the infidelity and sensitivity scores.
        The boxplots show the distributions of both evaluation metrics over the attributions of all explained samples, when different attribution methods are used. Our goal is to identify a method for which the attributions yield low sensitivity and infidelity.
        Our observations reveal that most methods exhibit high sensitivity to minor input perturbations, with only four methods maintaining flat scores when subjected to significant noise in the input data, i.e. flat infidelity scores. Notably, \gls{svs} stands out as the sole method that remains robust against both perturbation types.
        
\begin{figure}[h]
            \centering
            \includegraphics[width=0.9\columnwidth]{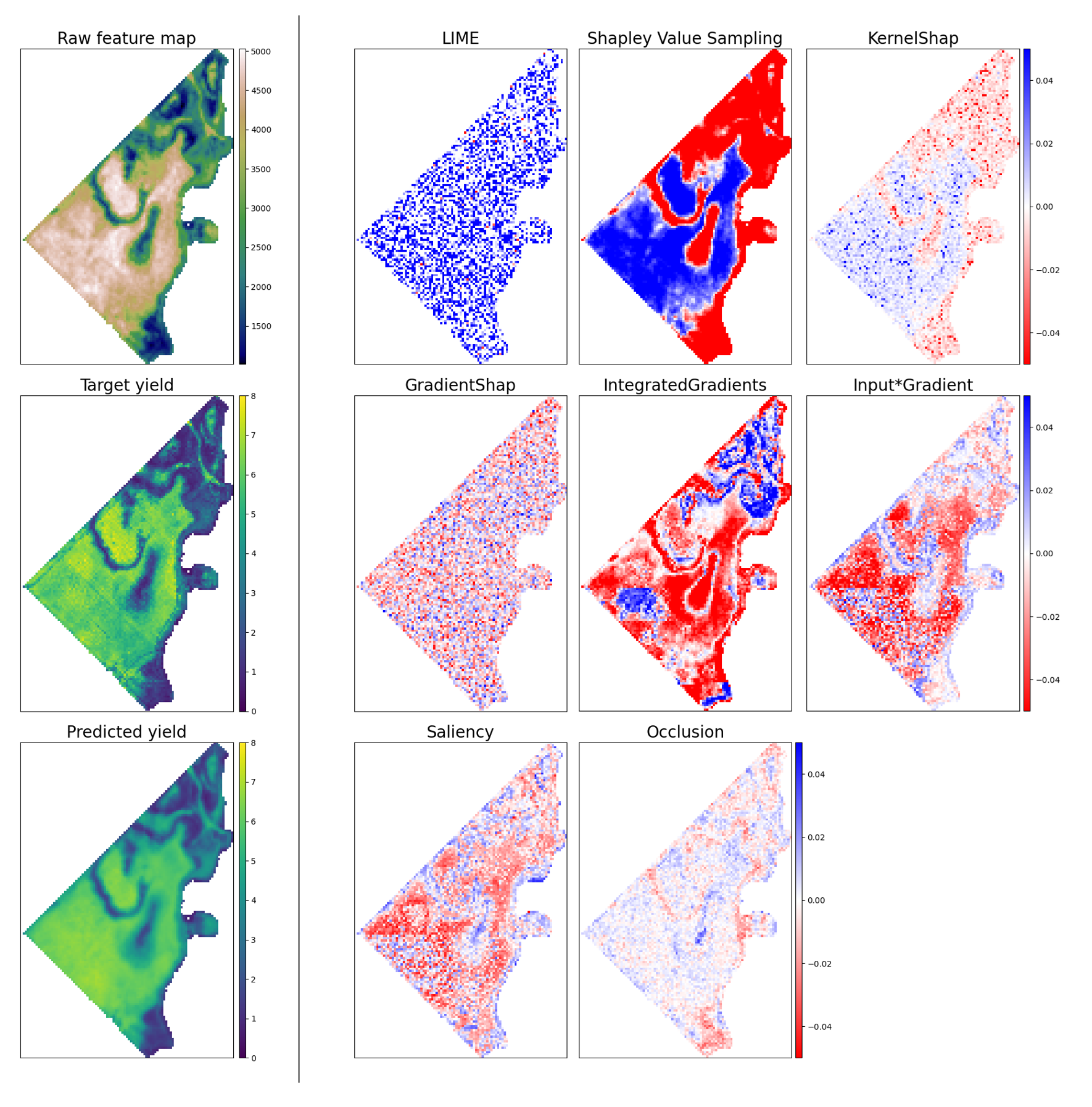}
            \caption{
                Qualitative evaluation of the feature attribution methods using the mean baseline on ARG-S dataset: Attribution maps of band B08 of time step 14 (March of the second year).
                }
            \label{fig:mean_baseline_eval_scores_a}
        \end{figure}

        We illustrate this characteristic further in Figure \ref{fig:mean_baseline_eval_scores_a} by examining the attribution maps of a single feature for various methods. 
        On the left column, the input values of the satellite band B08 on March of the second year are visualized on the first map, along with the reference and predicted yield maps of a field from ARG-S dataset. The remaining columns contain the attribution maps of the same feature and field when different methods are used.
        The blue and red values distinguish between positive and negative attributions, which have contributed to either increasing or decreasing the predicted yield value for each pixel, respectively. 
        We observe that only \gls{svs} produces a smooth attribution map, in contrast to the other attribution techniques where the results show pronounced spatial fluctuations. We verified that this pattern is consistent across the remaining features.
        Such behavior, indicative of high sensitivity, is undesirable because adjacent pixels often share similar input features and target values, implying that their attributions should also exhibit similarity \cite{alvarez2018towards,bhatt2020evaluating}.

\begin{figure*}[h]
            \centering
            \includegraphics[width=0.8\textwidth]{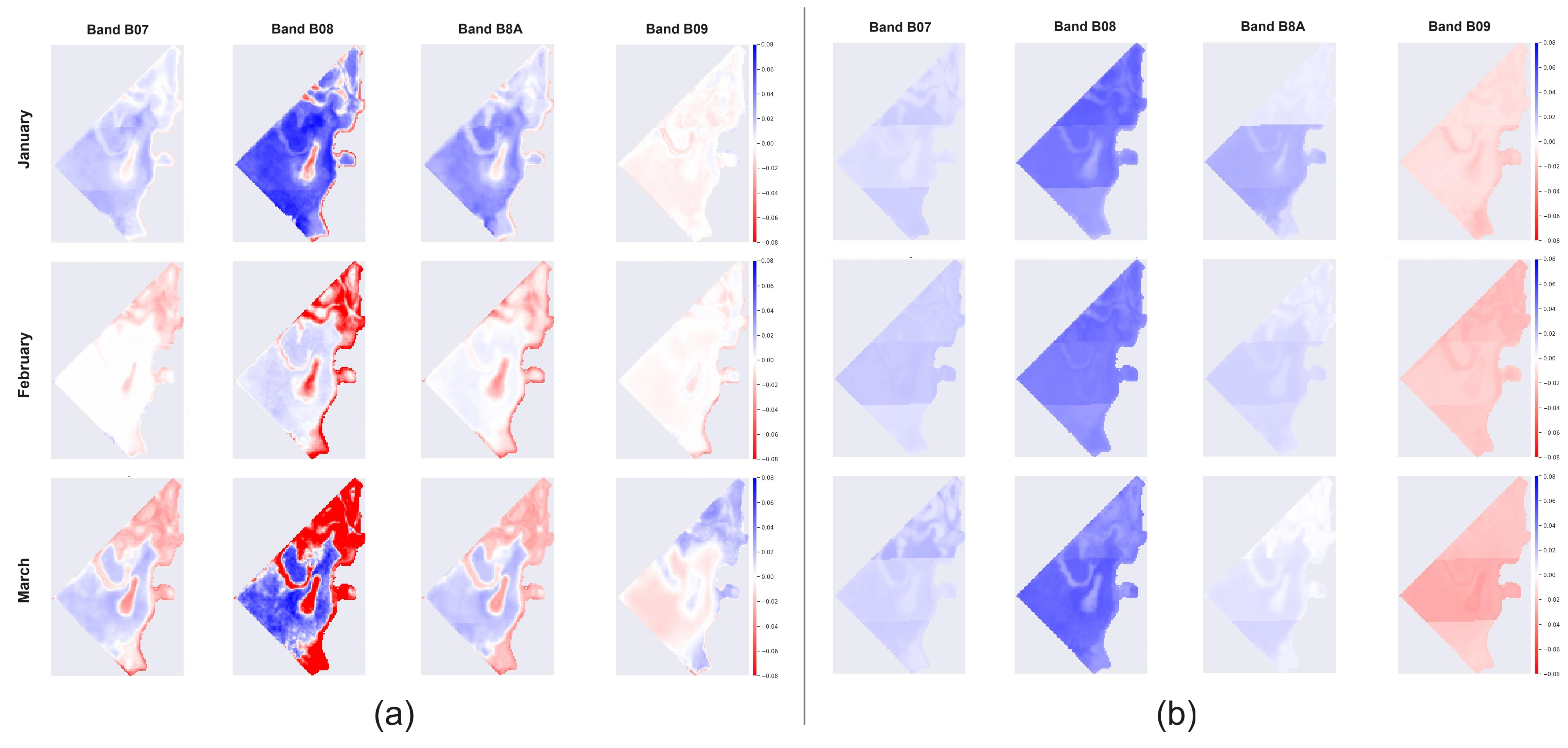}
            \caption{Comparing attribution maps of the \gls{svs} using the mean baseline (on the left) and the padded baseline (on the right), over three months from the second year and 4 bands satellite bands.}
            \label{fig:compare_baselines}
        \end{figure*}

        We further evaluate the choice of the baseline. While the results we described above are similar when using the padded baseline vector (see Appendix \ref{app:padded_baseline}), examining the attribution maps highlights a fundamental difference between the two options.
        In Figure \ref{fig:compare_baselines}, we select a field from the ARG-S dataset and visualize its attribution map for four satellite bands over three consecutive months. 
            In Figure \ref{fig:compare_baselines}.a, the results using the mean baseline effectively differentiate between high and low yield regions by displaying positive and negative attribution values, matching the variance on the target yield map (refer to Figure \ref{fig:mean_baseline_eval_scores_a}). 
            However, in Figure \ref{fig:compare_baselines}.b, the results of the padded baseline fail to capture similar patterns, with the maps consistently attributing values as either strictly positive or negative.
        
        We would like to finally highlight an important advantage of using this method, which stems from game theory; Shapley values inherently accounts for feature interactions by approximating the impact of a feature when added to all possible subsets of the remaining features \cite{lundberg2018consistent}. 

        Based on our analysis of the evaluation results and the above discussion, we conclude that \gls{svs} is a reliable interpretation method for feature importance estimation, and we therefore conduct the remaining experiments with the \gls{svs} attribution method, using the mean baseline.

    \subsection{Attribution analysis on different crops and regions}
    
        In this section, we investigate the interpretation results of the models trained on the different datasets and closely examine the spectral and temporal importance of each model. For each dataset, we examine the evaluation at each cross validation fold, and explain the model trained on the best-performing fold. We summarize the model evaluation results in Table \ref{table:best_cv_selection} in Appendix \ref{app:model}, with the selected folds highlighted in green.
        
        \subsubsection{Spectral Analysis}

\begin{figure*}[ht]
    \centering  
    \includegraphics[width=0.9\textwidth]{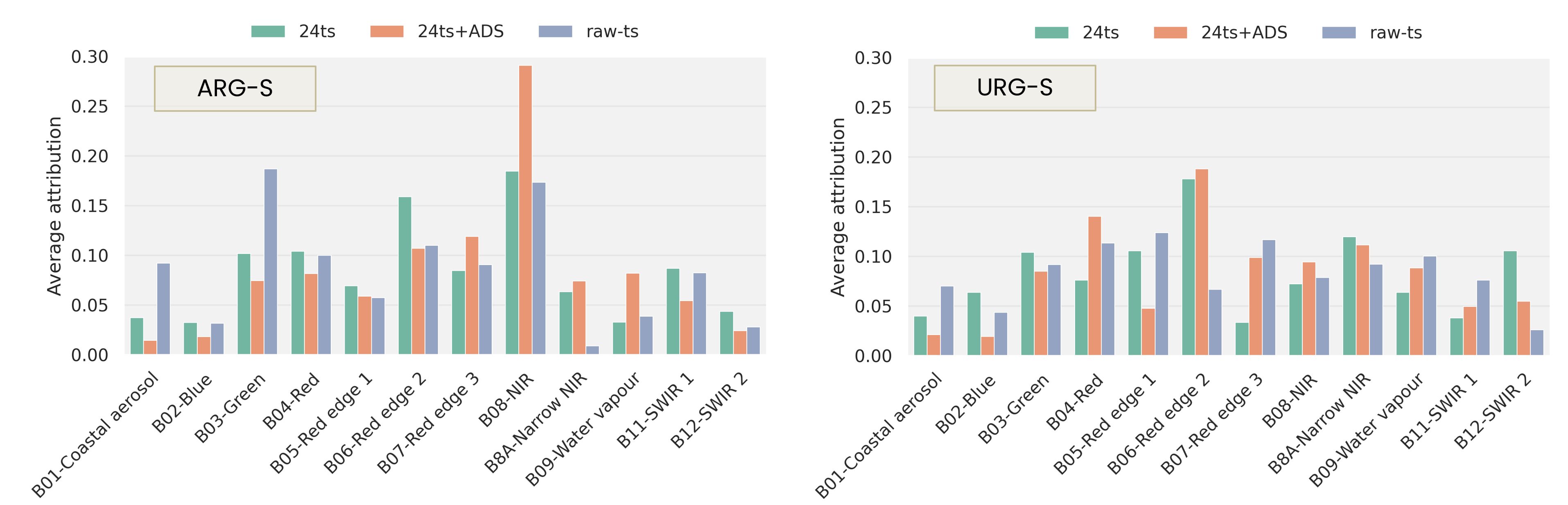}
    \includegraphics[width=0.9\textwidth]{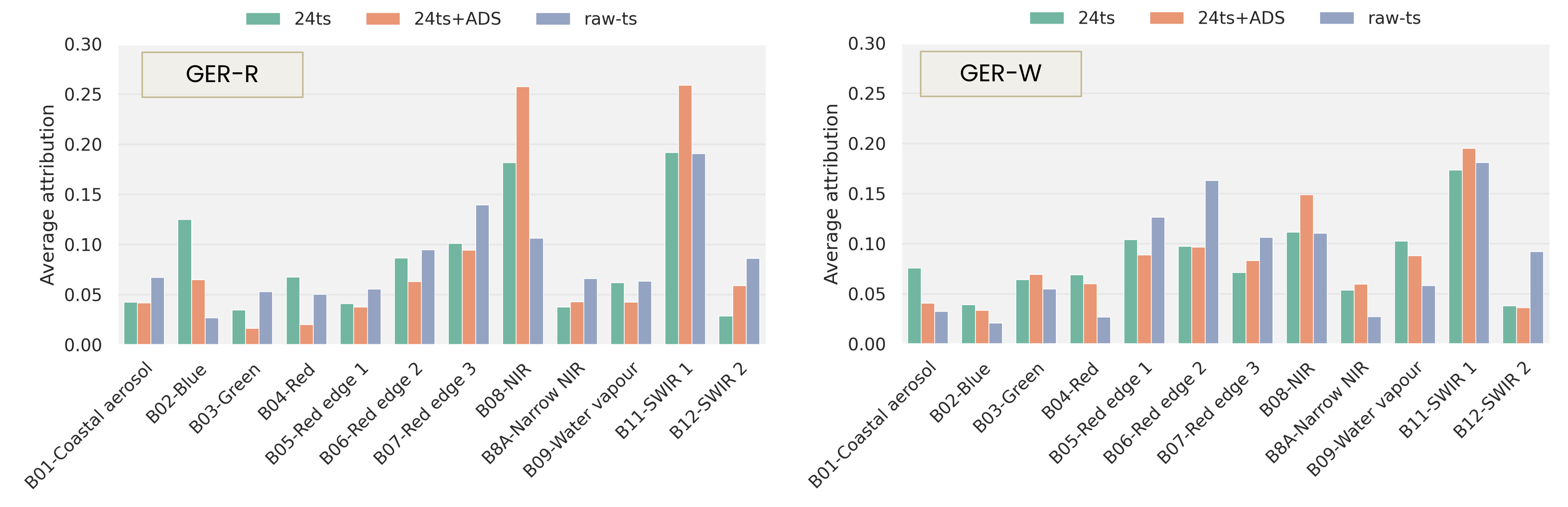}
    \caption{Average attributions per spectral band, in each dataset and under the three training settings: monthly sampling (\textit{24ts}), monthly sampling with additional data sources (\textit{24ts+ADS}), and raw time series (\textit{raw-ts}).
    }
    \label{fig:spectral_imp_bars}
\end{figure*}

            Across the three experiments conducted for each dataset, we compare the importance of satellite bands by evaluating the $\fct{SI}(X^i)$ for each pixel within the explained fields and then averaging the results. In experiments with additional modalities, we specifically rescale \gls{s2} attributions to ensure they sum up to 1, simplifying the comparison with the other two experiments. The results are illustrated in Figure \ref{fig:spectral_imp_bars}.

            When examining which bands are frequently attributed high importance across the different experiments, we observe that each band is significant in at least one experiment, meaning that the importances are distributed over all bands. 
            This explains why using only vegetation indices, which necessarily do not utilize all available bands, results in poorer performance compared to training with the full satellite spectrum, as previously discussed in \ref{subsec:res_yield_pred}.
            Our observation further provides an argument against restricting the satellite-based data to a set of indices or \gls{rgb} bands alone, a practice often carried out in other crop yield prediction studies  \cite{ wolaninEstimatingUnderstanding2020, dang2021autumn, nevavuori2020crop,  ma2021corn,jiang2020deep}.

            Comparing the raw against monthly temporal sampling strategies (i.e. \textit{raw-ts} and \textit{24ts}), we observe more similarities than differences between the corresponding bars across various bands and datasets. 
            In Table \ref{tab:spec_imp_similarity}, we compute in the first row the cosine similarity between these bars for each dataset.
            The values indicate that the band importances in Uruguay dataset are the most affected by the change in the temporal sampling strategy.

            \begin{table}[]
    \caption{Cosine similarity scores between the average spectral importances of different experiments.}
    \begin{tabular}{c|cccc}
        \textbf{Compared   experiments} & \textbf{ARG-S} & \textbf{URG-S} & \textbf{GER-R} & \textbf{GER-W} \\ \hline
        \textbf{24ts, raw-ts} & 0.93 & 0.84 & 0.90 & 0.93 \\
        \textbf{24ts, 24ts+ADS} & 0.92 & 0.91 & 0.95 & 0.98
    \end{tabular}
    \label{tab:spec_imp_similarity}
\end{table}
            
\begin{figure}[h]
                \centering  
                \includegraphics[width=0.99\columnwidth]{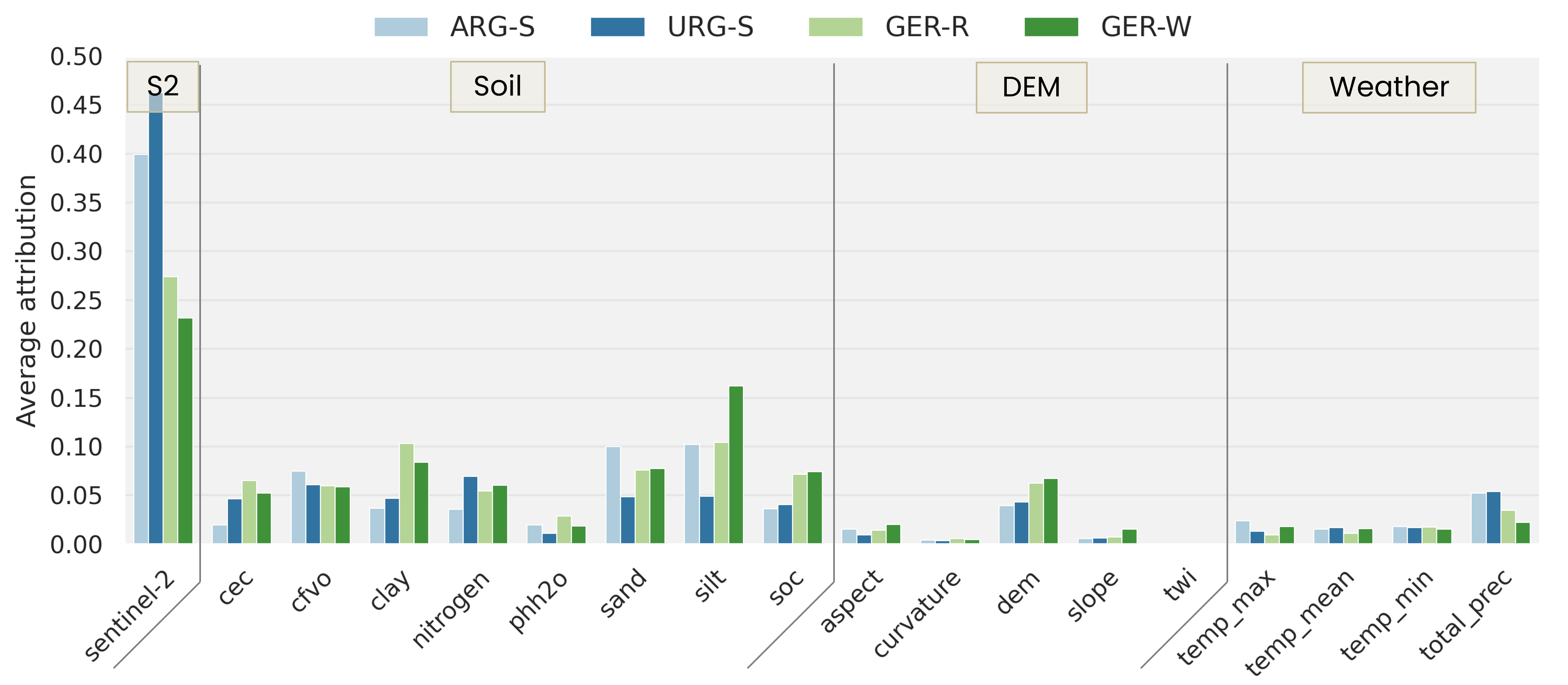}
                \caption{Average attribution of each feature in the \gls{ads}, contrasted with the total attribution of the satellite data across different datasets. }
                \label{fig:ADS_imp}
            \end{figure}

            To assess the impact of adding more modalities, we compare the monthly sampling experiments: \textit{24ts} and \textit{24ts+ADS}.
            We notice that in soybean datasets, i.e. in Argentina and Uruguay, the spectral importances between the two experiments are less similar than in rapeseed and wheat datasets in Germany, as reported in the second row of Table \ref{tab:spec_imp_similarity}.
            We further examine the total attribution of satellite data relative to each feature from the additional modalities, as depicted in Figure \ref{fig:ADS_imp}.
            We notice that while \gls{s2} accounts for nearly half of the attributions in Argentina, and around 40\% in Uruguay, it contributes only about 25\% of the total attribution in rapeseed and wheat datasets. 
            The majority of the remaining attribution is distributed among the soil bands, followed by weather and terrain elevation features.
            In the case of the \gls{dem} data, we notice that most of the attribution is concentrated on the raw elevation values (i.e. the \textit{dem} feature), suggesting that the additional modalities can be further filtered to use only the features relevant for the yield prediction.
            
\begin{figure}[h]
                \centering  
                \includegraphics[width=0.99\columnwidth]{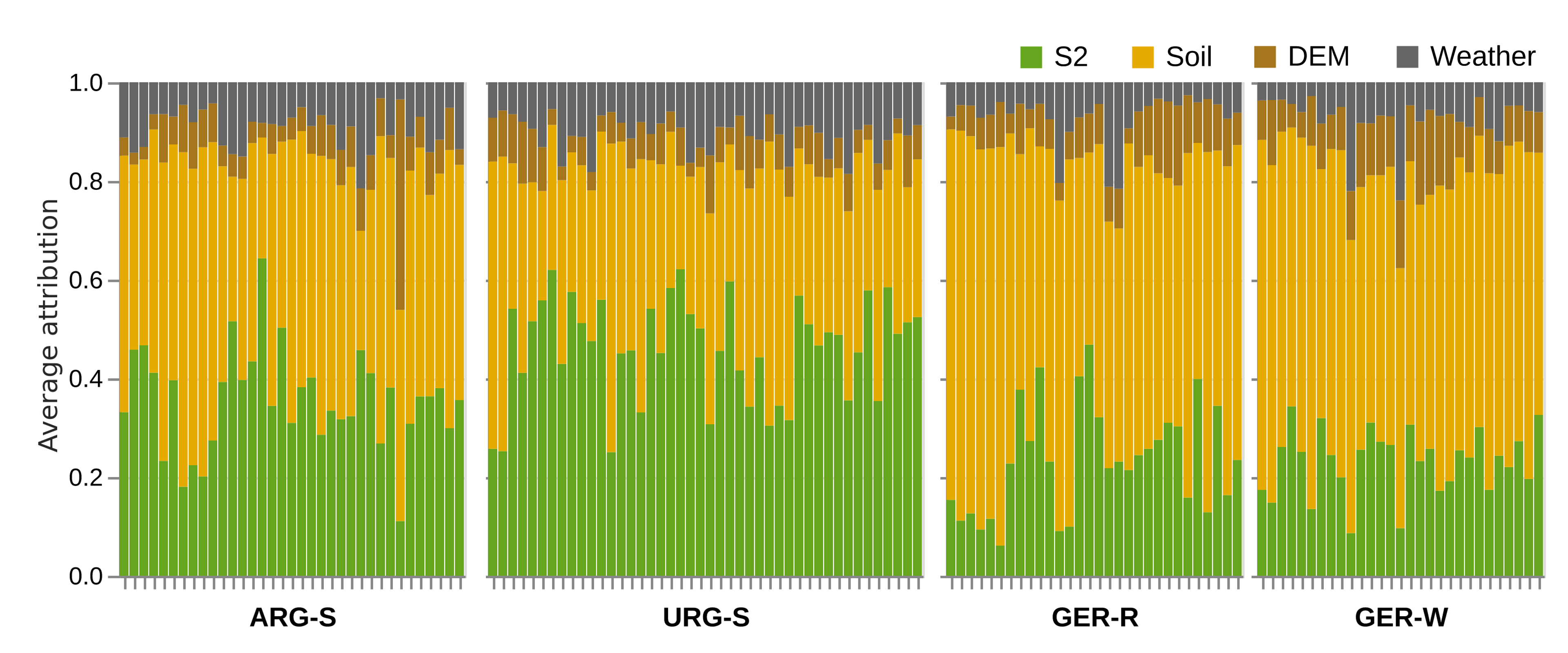}
                \caption{Total modality attribution, averaged over pixels of each explained field separately (i.e. each bar is a field), in each dataset.}
                \label{fig:view_attr}
            \end{figure}

            We further compare the total attribution per data source in Figure \ref{fig:view_attr}, differentiating between the explained fields (i.e., the vertical bars) in each dataset. 
            The attributions are averaged across the pixels of each field separately.
            The area that each color covers serves as a visual indicator of the total importance each modality has in the different datasets. 
            The yellow color, in particular, highlights the significance of soil bands in most fields and datasets, particularly in Germany, where it substantially surpasses the total attribution of \gls{s2} bands.
            In contrast, weather and \gls{dem} exhibit relatively lower importance across all regions, with weather being more important for soybean crops, and \gls{dem} showing more significance for rapeseed and wheat crops in Germany.  
            It is worth noting that the literature, particularly the work of Thomson et al. \cite{thomson2002elevation}, emphasizes the important impact of topography on wheat crop productivity, further supporting our findings.
            
        \subsubsection{Temporal Analysis}

\begin{figure*}[h]
                \centering
                \includegraphics[width=0.88\textwidth]{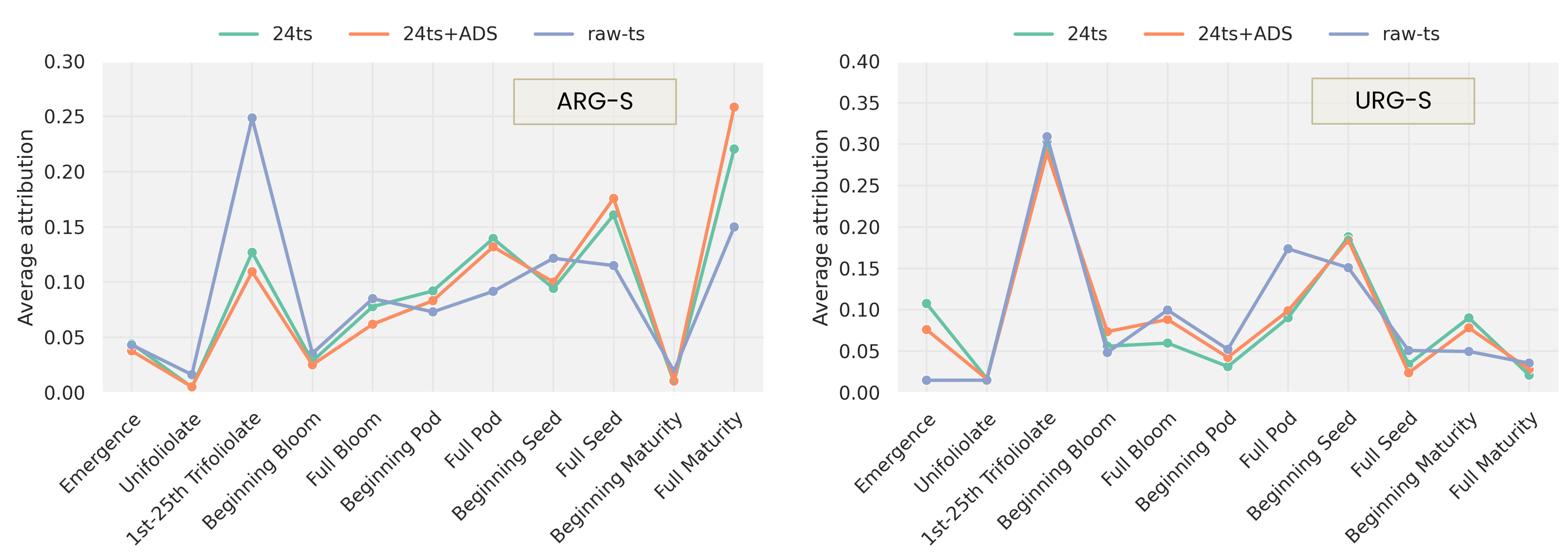}
                \includegraphics[width=0.90\textwidth]{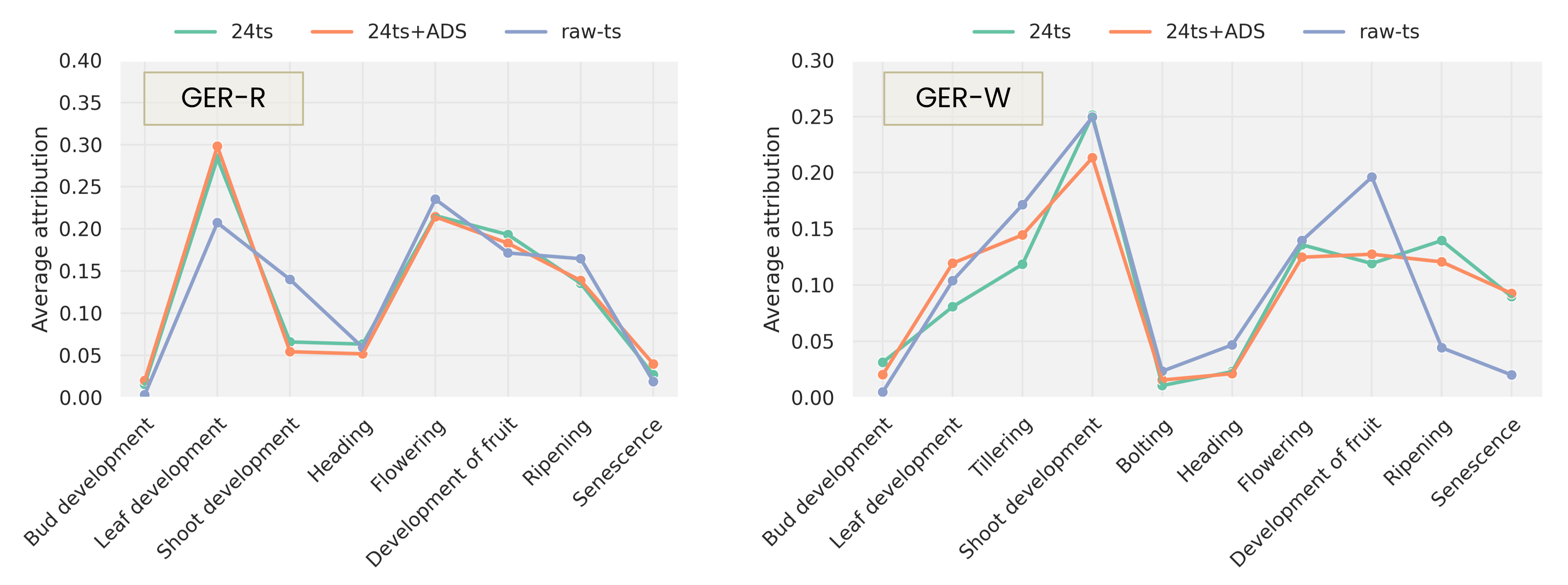}
                \caption{Average of total attributions per growth stage in each dataset, under the three training settings: monthly sampling (\textit{24ts}), monthly sampling with additional data sources (\textit{24ts+ADS}), and raw time series (\textit{raw-ts}).}
                \label{fig:temp_imp_bars_GS}
            \end{figure*}

            To analyze the temporal attribution results, we directly assess their aggregations over the growth stages, $\fct{TI}(X^i)_{t_{\text{gs}}}$, averaged across all the explained data points $X^i$. We plot the results for the different experiments and datasets in Figure \ref{fig:temp_imp_bars_GS}, and report their corresponding similarity scores in Table \ref{tab:temp_imp_similarity}.
            
            \begin{table}[]
    \caption{Cosine similarity scores between the average temporal importances aggregated along growth stages of different experiments.}
    \begin{tabular}{c|cccc}
        \textbf{Compared   experiments} & \textbf{ARG-S} & \textbf{URG-S} & \textbf{GER-R} & \textbf{GER-W} \\ \hline
        \textbf{24ts, raw-ts} & 0.91 & 0.94 & 0.96 & 0.92 \\
        \textbf{24ts, 24ts+ADS} & 0.99 & 0.99 & 1.00 & 0.99
    \end{tabular}
    \label{tab:temp_imp_similarity}
\end{table}

            We observe that the two experiments utilizing monthly temporal sampling, \textit{24ts} and \textit{24ts+ADS}, treat data from various stages similarly, since the two corresponding curves closely align across different datasets. This is further confirmed by the similarity scores higher than 0.99\%, indicating that training with additional modalities have minimal impact on the significance of each growth stage to the final predictions. 
            However, when we compare the \textit{24ts} and \textit{raw-ts} curves, we observe a more disparities in their values at many stages. 
            We therefore conclude that utilizing the raw time series of satellite data influences the importance of each growth stage in the model. 
            This outcome aligns with our expectations, as the distribution of growth stages over time is more finely-grained than the monthly sampling, making the density of the raw time series better suited to adequately capture each growth stage.
            Among the stages which gain more importance when using the full \gls{s2} time series are 
                (i) the beginning of bloom, pod and maturity in ARG-S,
                (ii) full blooming and beginning and full pod in URG-S,
                (iii) shoot development and heading in GER-R,
                (iv) and finally bolting, heading and flowering in GER-W.

\begin{figure}[h]
                \centering  
                \includegraphics[width=0.99\columnwidth]{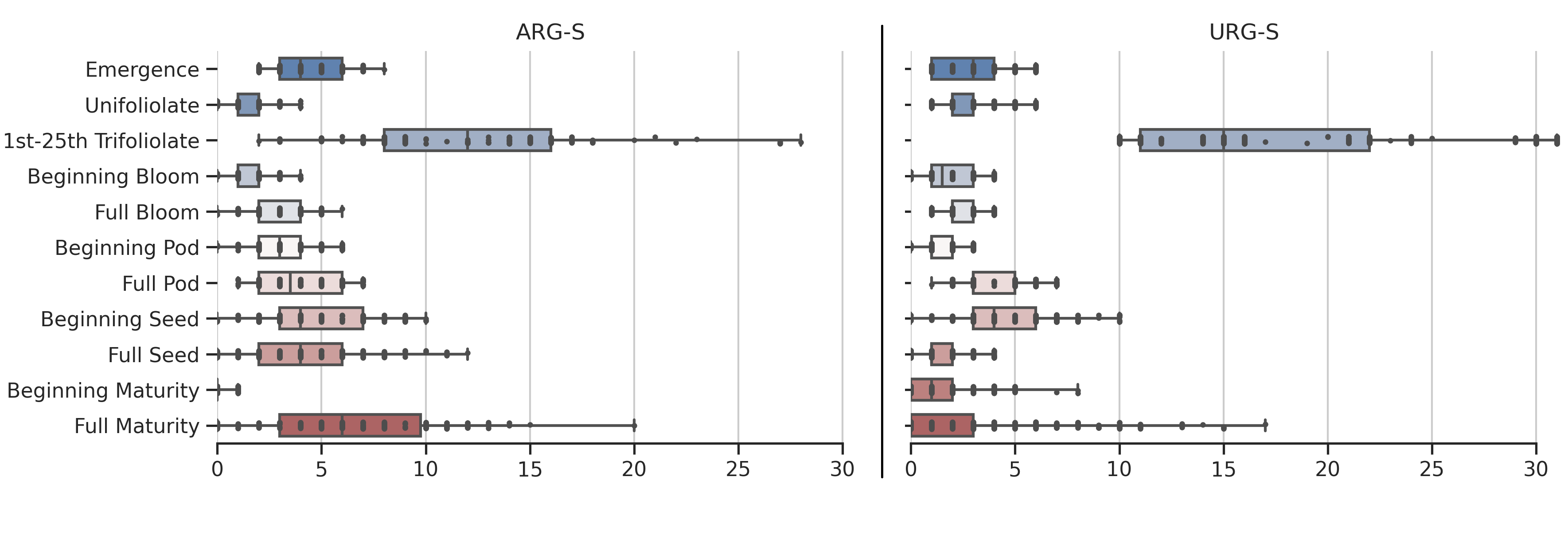}
                \includegraphics[width=0.99\columnwidth]{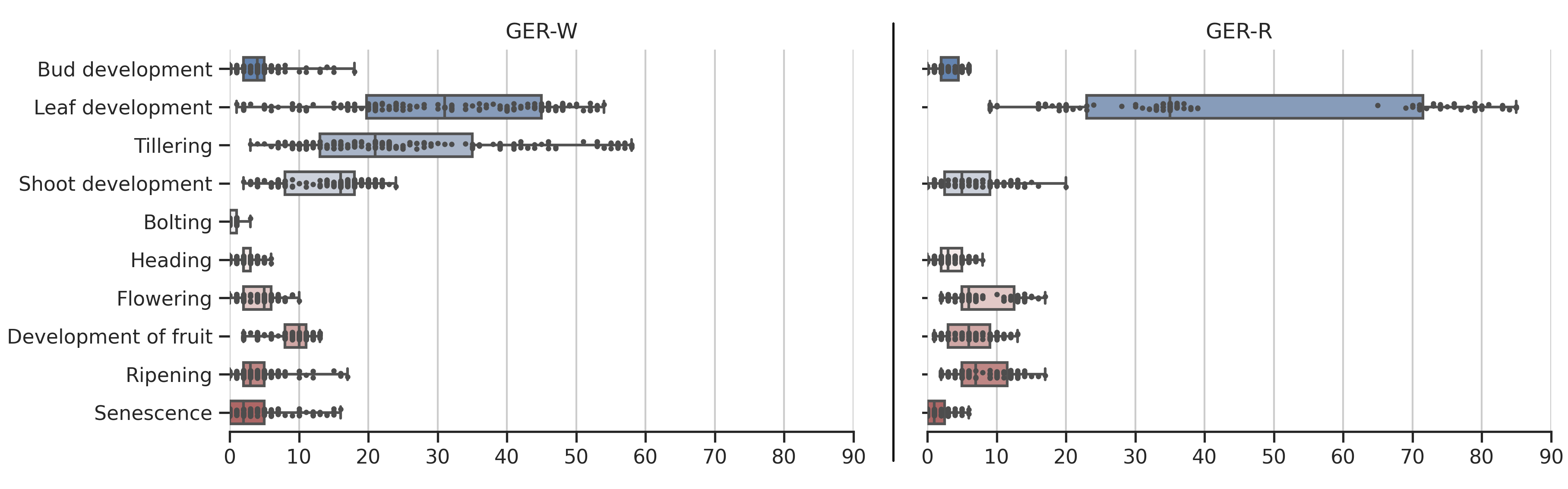}
                \caption{Number of satellite instances available per growth stage in each field of each dataset.}
                \label{fig:data_per_gs}
            \end{figure}

            To assess the impact of growth period lengths on the attribution of experiments utilizing raw time series data, we count the number of satellite instances in each stage and for each field in the different datasets, and we plot the results in Figure \ref{fig:data_per_gs}.
            Upon examination of the soybean datasets, we notice that the extended duration of the third stage and the short late stages positively correlate with their respective attributions in the \textit{raw-ts} experiment.
            In the rapeseed dataset, a similar pattern is observed for the early stages, while the late stages have a high importance despite their short period.
            In the case of the wheat dataset, the correlation is weaker, where stages such as shoot and fruit development, despite not being the longest in duration, exhibit the higher attributions. 
            These observations lead us to conclude that while in soybean datasets, the length of growth stages and the abundance of corresponding satellite data positively correlate, such a relationship has less significance in wheat and rapeseed crops.

            Our analysis suggests that while the duration of growth stages may impact their influence on the model, it is not the sole factor influencing their overall attribution.
            Hence, it is reasonable to assume that the model has captured some crop-specific characteristics rooted in agrobiology and phenology. These characteristics, in turn, shape its predictions, resulting in a closer alignment with actual yield factors. We further validate this assumption in the discussion in section \ref{sec:discusison}.

    \subsection{Further analysis}
    
        \subsubsection{In-Field variability}

            We closely examine the attribution maps to gain insight into how the model captures in-field variability. 
            This is significant because our approach provides the model with individual pixel data without any information about neighboring pixels, yet the model learns to approximate the variability we observe in the reference yield maps.
            When revisiting the attribution maps in Figure \ref{fig:compare_baselines}.a, we notice that some features have high variance across the field's pixels, while others consistently show either low or high attribution values.
            
\begin{figure}[h]
    \centering
    \includegraphics[width=\columnwidth]{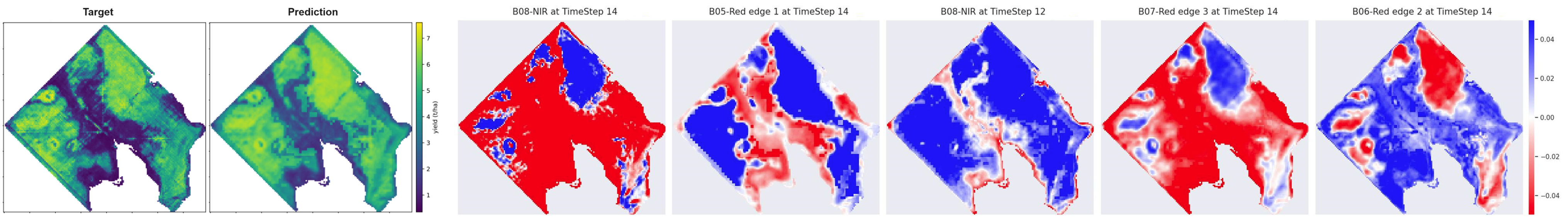}
    \vspace{2pt} 
    \includegraphics[width=\columnwidth]{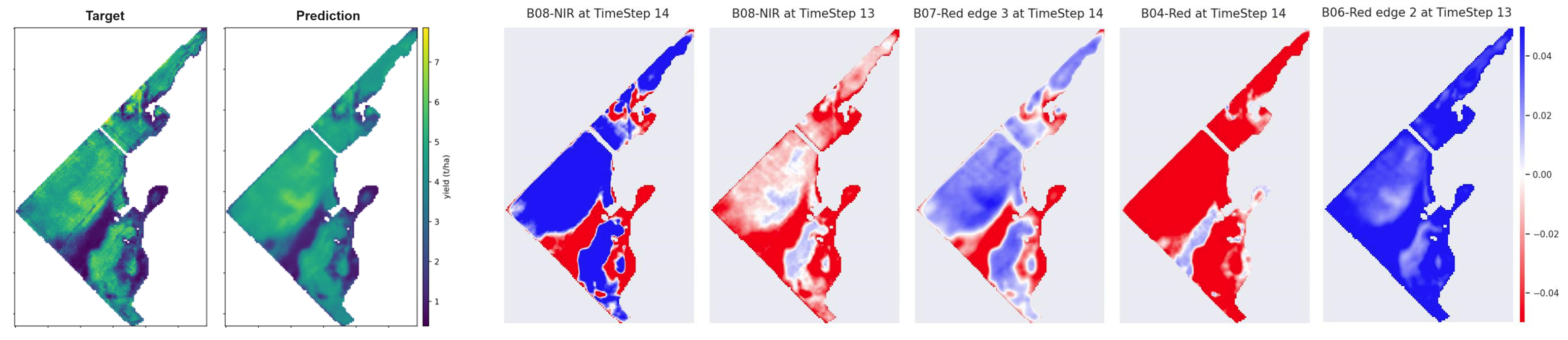}
    \vspace{2pt} 
    \includegraphics[width=\columnwidth]{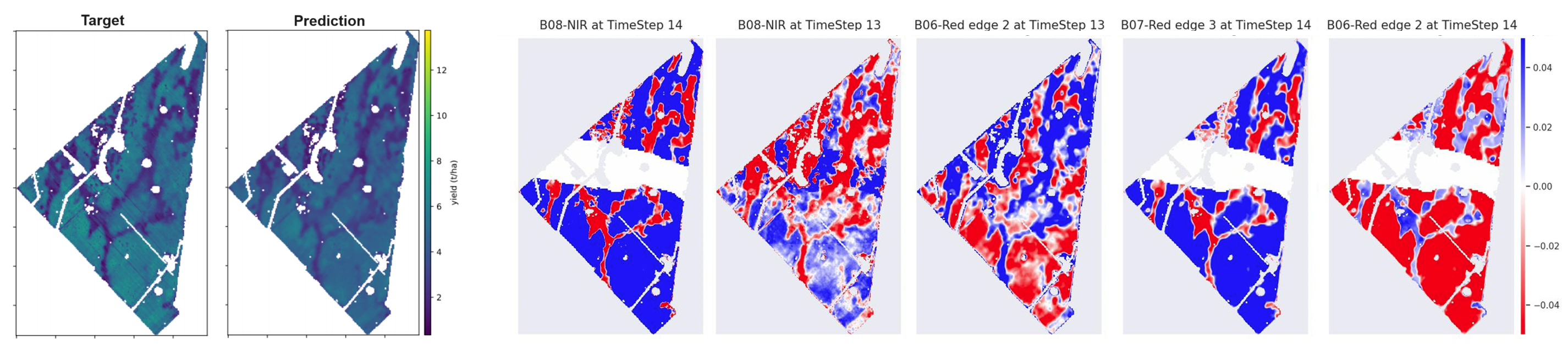}
    \caption{In-field variability of 3 fields from ARG-S. From left to right: target and predicted yield maps, followed by the attribution maps of five features with the highest attribution variance. (The horizontal strip on some maps from the field at the bottom are padded pixels because of cloud coverage.)}
    \label{fig:field_attr_var}
\end{figure}

            To identify band and time step pairs that effectively capture the in-field yield variability learned by the model, we calculate the variance across different pixels for each feature and each field. In Figure \ref{fig:field_attr_var}, we present the results for three fields in Argentina by displaying the 5 features with the highest attribution map variance.
            We observe that the patterns on the attribution maps match those in the prediction map, confirming that features with high variability are key in helping the model best capture the yield variability on the subfield level.
                
\begin{figure}[h]
                \centering  
                \includegraphics[width=0.99\columnwidth]{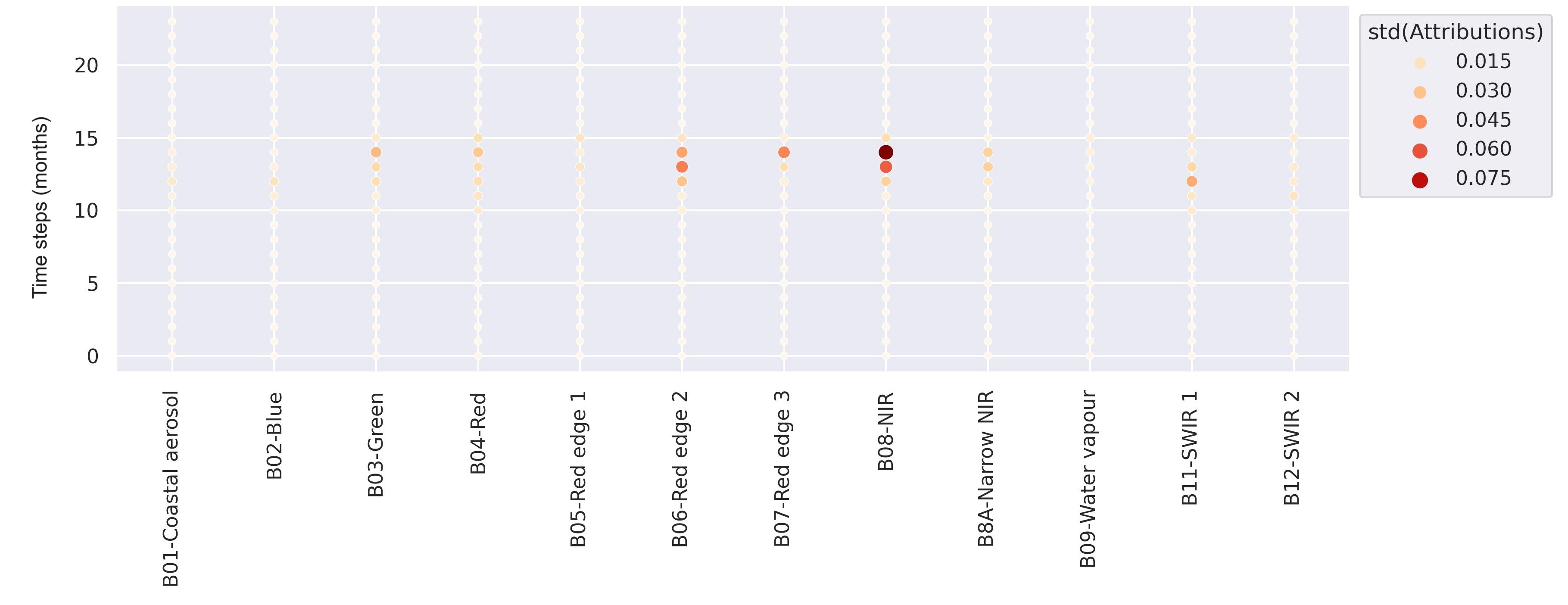}
                \caption{
                Standard deviation of the relative scaled attribution values across all interpreted pixels from a single field in Argentina dataset.}
                \label{fig:arg_attr_grid_std}
            \end{figure}

            The grid of the standard deviations of the relative scaled attributions (i.e. $\mathbf{\hat{a}^i}$) shown in Figure \ref{fig:arg_attr_grid_std}, can be used to directly identify the features that have a significant influence on the model's ability to learn the in-field variability for a specific field. In this example, we computed the attribution deviations for the field depicted at the bottom of Figure \ref{fig:field_attr_var}.
            In the scenario of training with monthly sampling, as illustrated in the figure, these influential features are primarily the NIR and Red Edge 2 bands during the last months before harvesting. The harvesting season typically occurs around time step 15, i.e., in April of the second year, in most fields.
            
        \subsubsection{Attribution and performance correlation}

\begin{figure}[h]
                \centering  
                \includegraphics[width=0.99\columnwidth]{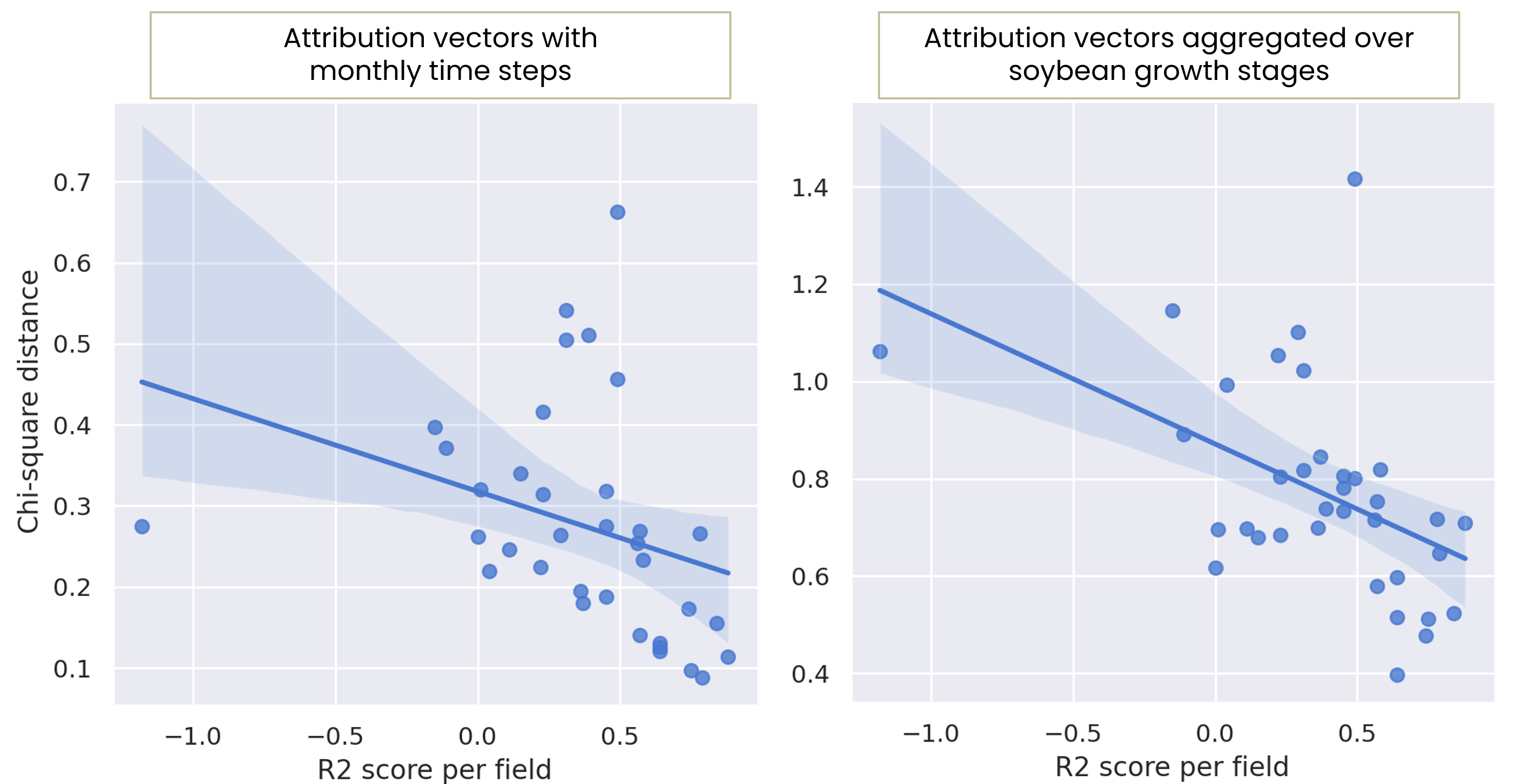}
                \caption{
                Comparison of the \gls{r2} score of each field against the distance of the attribution vectors of its pixels to the reference distribution. Each point is a field in ARG-S. Here, the reference distribution is the average distribution over all explained data points in Argentina.
                }
                \label{fig:scatterplot_r2score_attr_distance}
            \end{figure}

            When evaluating the model on different fields, we observe significant variations in the results. Our goal is to determine if a  field for which the model performs poorly has an attribution distribution that differs from fields where the model performs better.
            To achieve this, we consider two types of attribution vectors: the original vector, which contains 24 time steps, or its aggregation over the growth stages.
            For each type, we start by defining the reference vector as the average across all considered data points. Next, we calculate the distance between the average attributions of each field and this reference vector.
            The results of this analysis are shown in Figure \ref{fig:scatterplot_r2score_attr_distance}, with a focus on the ARG-S dataset.

            We observe a negative correlation with both attribution vector types, indicating that fields with a high \gls{r2} score tend to have a closer distance to the reference distribution. 
            To further quantify this correlation and its statistical significance, we conduct a t-test with a threshold of 5\%. The results are reported in Table \ref{table:corr_tests_arg24ts} (see the rows marked with \textit{(all fields)}). 
            The p-value exceeds 6\% when using the monthly attribution vectors, while it falls below 1\% with the vectors aggregated over growth stages. Consequently, the latter method is more reliable in identifying fields with low performance.

\begin{table}[h]
\footnotesize
\centering
\caption{Correlation and t-test results between the attribution distance of ARG-S explained fields to the different references, and their \gls{r2} score. The attribution results are of the model trained under the monthly sampling. P-values below 0.05 are in bold. }
\begin{tabular}{|c|c|c|c|c|}
\hline
\begin{tabular}[c]{@{}c@{}}\textbf{Temporal}\\ \textbf{dimension}\end{tabular} & \begin{tabular}[c]{@{}c@{}}\textbf{Score}\\ \textbf{threshold}\end{tabular} & \begin{tabular}[c]{@{}c@{}}\textbf{Best}\\ \textbf{N fields}\end{tabular} & \textbf{Correlation} & \textbf{P-value} \\ \hline
\multirow{8}{*}{\begin{tabular}[c]{@{}c@{}}Monthly\\ time steps\end{tabular}} & - & (all fields) & -0.32 & 0.063 \\
 & -  & 3 & -0.30 & 0.076 \\
 & -  & 5 & -0.33 & 0.056 \\
 & -  & 10 & -0.38 & \textbf{0.025} \\
 & -  & 15 & -0.39 & \textbf{0.021} \\
 & 0.4 & -  & -0.39 & \textbf{0.020} \\
 & 0.5 & -  & -0.38 & \textbf{0.024} \\
 & 0.6 & -  & -0.38 & \textbf{0.023} \\ \hline
\multirow{8}{*}{\begin{tabular}[c]{@{}c@{}}Aggregated\\ over growth \\ stages\end{tabular}} & - & (all fields)  & -0,48 & \textbf{0,003} \\
 & - & 3 & -0,30 & 0,077 \\
 & - & 5 & -0,38 & \textbf{0,025} \\
 & - & 10 & -0,43 & \textbf{0,009} \\
 & - & 15 & -0,46 & \textbf{0,005} \\
 & 0,4 & - & -0,48 & \textbf{0,004} \\
 & 0,5 & - & -0,45 & \textbf{0,007} \\
 & 0,6 & - & -0,44 & \textbf{0,008} \\ \hline
\end{tabular}
\label{table:corr_tests_arg24ts}
\end{table}

            We further explore different approaches to define the reference vector. One reliable baseline is obtained by averaging the attributions across the best-performing fields. These fields can be identified either by setting a threshold on their \gls{r2} score or by specifying a predetermined number of fields with the highest \gls{r2} scores.
            The corresponding experiments are listed in the second and third columns (i.e. \textit{Score threshold} and \textit{Best N fields}) in Table \ref{table:corr_tests_arg24ts}, respectively. We highlight the p-values below 5\% are in bold. 
            We observe that most of the tests have yielded a significant correlation, especially when using the aggregated attribution vector, which further supports its superior efficiency over the original vector in distinguishing between the fields on which the model is well- or poorly-performing.

            While the results described above originate from the model trained under the monthly sampling setting, we repeat the same experiments for the model trained on the raw time series, and report the corresponding results in Table \ref{table:corr_tests_arg_rawts}. 
            We notice in this case that no test exhibits a statistically significant correlation when using the raw time series. 
            Conversely, when comparing growth stage aggregations, we notice a contrary trend. We attribute this behavior to the varying sequence lengths resulting from the utilization of raw time series data, where each time step originates from different timestamps across various fields.
            In contrast, comparing the importances of growth stages proves to be more robust against these temporal disparities.

\begin{table}[h]
\centering
\footnotesize
\caption{Correlation and t-test results between the attribution distance of ARG-S explained fields to the different references, and their \gls{r2} score. The attribution results are of the model trained with the raw satellite time series.}
\begin{tabular}{|c|c|c|c|c|}
\hline
\begin{tabular}[c]{@{}c@{}}\textbf{Temporal}\\ \textbf{dimension}\end{tabular} & \begin{tabular}[c]{@{}c@{}}\textbf{Score}\\ \textbf{threshold}\end{tabular} & \begin{tabular}[c]{@{}c@{}}\textbf{Best}\\ \textbf{N fields}\end{tabular} & \textbf{Correlation} & \textbf{P-value} \\ \hline
\multirow{8}{*}{\begin{tabular}[c]{@{}c@{}}raw\\ time series\end{tabular}} & - & (all fields) & -0.17 & 0.324 \\
& - & 3 & -0.27 & 0.122 \\
 & - & 5 & -0.22 & 0.205 \\
 & - & 10 & -0.17 & 0.326 \\
 & - & 15 & -0.17 & 0.336 \\
 & 0.4 & - & -0.19 & 0.267 \\
 & 0.5 & - & -0.18 & 0.291 \\
 & 0.6 & - & -0.18 & 0.296 \\ \hline
\multirow{8}{*}{\begin{tabular}[c]{@{}c@{}}Aggregated\\ over growth \\ stages\end{tabular}} & - & (all fields) & -0,37 & \textbf{0,0286} \\
 & - & 3 & -0,34 & \textbf{0,0490} \\
 & - & 5 & -0,36 & \textbf{0,0358} \\
 & - & 10 & -0,39 & \textbf{0,0202} \\
 & - & 15 & -0,39 & \textbf{0,0220} \\
 & 0,4 & - & -0,38 & \textbf{0,0225} \\
 & 0,5 & - & -0,38 & \textbf{0,0232} \\
 & 0,6 & - & -0,38 & \textbf{0,0236} \\ \hline
\end{tabular}
\label{table:corr_tests_arg_rawts}
\end{table}

        \subsubsection{Soybean data analysis}
            We further analyze the attribution results for the fields from the Argentina and Uruguay datasets. We anticipate similar results for both dataset, as they contain the same crop. Additionally, due to the geographic proximity of the two countries, we expect a high degree of similarity in the analysis outcomes. In order to analyze these differences, we make use of dimensionality reduction tools to verify the similarity between the input data and attribution results from both countries. 

            We randomly select and combine 10,000 points from each dataset and compare two vectors from each sample: the input satellite data and the attribution vector. Furthermore, we aggregate the attributions over the growth stages, as the raw vectors may contain field-specific patterns related to the off-season padded time steps. We then project them into a two-dimensional embedding space using \gls{pca} and present the results in Figure \ref{fig:dim_reduction_pca}. The corresponding T-SNE plots are also attached in Appendix \ref{app:dim_reduction_tsne}.

            \begin{figure*}[h]
                \centering  
                \includegraphics[width=0.95\textwidth]{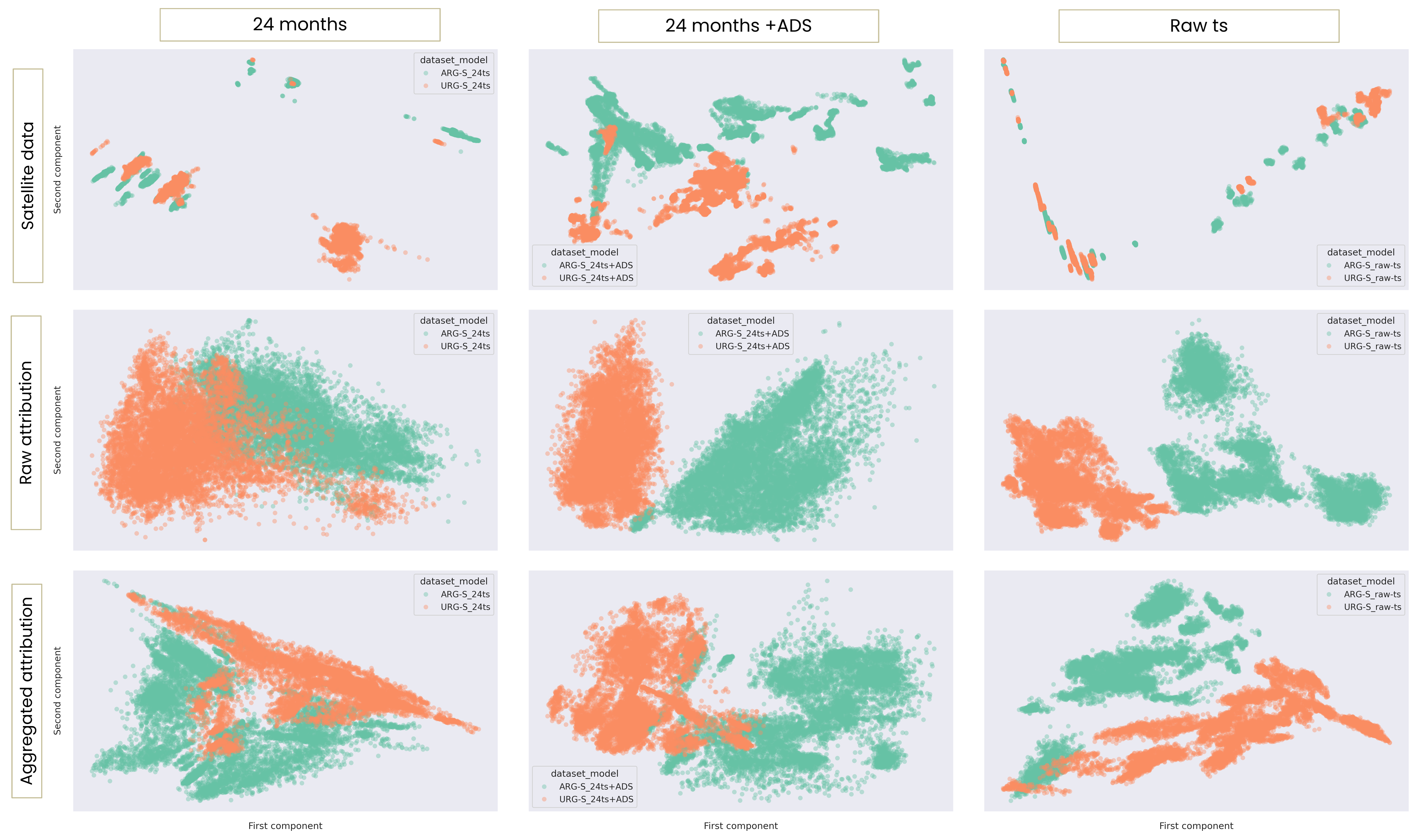}
                \caption{\gls{pca} results comparing Argentina and Uruguay data. The columns correspond to the different training settings, and the rows correspond to the type of vectors on which the dimensionality reduction is applied.}
                \label{fig:dim_reduction_pca}
            \end{figure*}

            Contrary to our expectations, our results indicate that the spectral and temporal importance show distinct patterns between the two datasets.
            We first examine the satellite vectors. We can observe that data from the same country consistently forms multiple clusters in all of our experiments. This suggests that the source data from Argentina inherently differs from that of Uruguay. Additionally, we can see that the two datasets are much more distinguishable when we include additional modalities in our training, as opposed to using only \gls{s2} data. This observation implies that information related to soil, weather, and terrain elevation exhibits distinct patterns in the two countries.

            Regarding the attribution vectors, on the second and third row in Figure \ref{fig:dim_reduction_pca}, we observe a clear separation between Argentina and Uruguay data points in the case of training with the full raw time series. However, in the monthly sampling experiments, there is a slight overlap between these data points. 
            These results imply that incorporating more time steps into the satellite time series leads to the emergence of more region-specific characteristics within the data. 
            When comparing the raw attribution vectors with their aggregation over growth stages, we note that the latter exhibits greater overlap between the two datasets. 
            We assume that this discrepancy may be attributed to the issue of time series misalignment in the raw data.

\section{Discussion}\label{sec:discusison}
    
    \subsection{Spectral importance and Vegetation Indices}
        In the following, we connect the spectral importance results and the model performance when trained on vegetation indices.  
        The \gls{ndvi} is frequently used for yield prediction as highlighted in \cite{murugananthamSystematicLiterature2022}, and relies on the near-infrared (B08) and red (B04) bands.
        In the spectral importance results in Figure \ref{fig:spectral_imp_bars}, we notice that B08 emerges as significantly important in all datasets, aligning with its universal relevance, while B04 is assigned high attributions in soybean datasets in particular. 
        Additionally, the Sentinel-2 satellite imagery incorporates three red edge bands, of which only two are actually employed in the vegetation indices we utilized, as can be verified in Table \ref{tab:VIs}. 
        However, the results underline that the third red edge band (B07) also exerts a high influence on the models across all experiments. 
        Moreover, the first SWIR band (B11), also absent from the indices we used, has particularly high attributions in rapeseed and wheat datasets. This can potentially explain the substantial drop in the VI experiments for these two datasets, as reported in Table \ref{table:model_raw_ts_eval} and \ref{table:model_monthly_eval}.

        The spectral importance enables the identification of the most important bands for each crop and region, and these results can efficiently be exploited to define specialized indices for the crop yield prediction task. Combining these results with the growth stage attributions, the vegetation indices can further be exploited to estimate the length of each growth stage.
    
    \subsection{Explaining temporal importances through domain knowledge}
        In this section, we aim to compare our temporal importance results against well-established facts in crop growth and development research. 
    
        In soybean datasets, \textbf{emergence}, \textbf{unifoliolate} and \textbf{trifoliolate} stages collectively constitute the vegetation phase. The main stem nodes and their branching develop during this phase, influencing the canopy structure and the final number of nodes, as elucidated by Kumudini \cite{kumudini2010soybean}.
        Since various environmental and genetic factors can impact the vegetative development, a poor canopy expansion can be identified at the last stage (i.e. \textbf{1st-25th trifoliolate}) and linked by the model to lower yield values.
        The \textbf{beginning pod} stage is also strongly related to yield, with favorable temperature and moisture conditions resulting in a higher pod number per plant, more beans per pod, and larger seeds \cite{mcwilliams1999soybean}.
        The following three stages, which extends from \textbf{full pod} to \textbf{full seed} stages, are important as the pods grow rapidly, and seed development begins, making it a crucial period for seed yield \cite{mcwilliams1999soybean}.
        This seed-filling phase is proven to have a positive correlation with yield \cite{gay1980physiological,smith1986selection}, and any stress during this phase has a more significant impact on yield reduction than at other times\cite{mcwilliams1999soybean}. 
        Finally, the pronounced significance of the full maturity stage in Argentina can be attributed to the enhanced visibility of the pods. At this stage, approximately 95\% of the pods on the main stem attain their mature pod color, offering a valuable visual indicator of yield, as supported by Kumidini \cite{kumudini2010soybean}.
    
        One of the factors that could account for the differences in the results between Argentina and Uruguay is the notable influence of the flowering stage. As documented in \cite{mcwilliams1999soybean, kumudini2010soybean}, soybean flowering is influenced by factors such as crop variety, day length, and temperature. Consequently, the variations observed in these regions may be attributed to differences in agricultural practices, including the application of fertilizers and herbicides.
    
        In rapeseed, the attribution results of juvenile growth (i.e. from \textbf{leaf development} to \textbf{heading}) is attributed to the close connection between the seed yield and the number of pods per plant, which is supported by the dry matter produced during this period. There is a linear relationship between the cumulative production of dry matter until flowering and pod density, as documented by Diepenbrock \cite{diepenbrock2000yield}.
        He further highlights that the \textbf{flowering stage} is the most pivotal phase impacting rapeseed yield, which aligns with the high attribution of this stage in the bottom left plot of Figure \ref{fig:temp_imp_bars_GS}. The ability of satellite data to capture this crucial stage is attributable to the increasing flower cover, which significantly enhances photon reflectivity and absorption to 60-65\% of incoming radiation.
        At the \textbf{fruit development stage}, pod filling is initiated and continues through the \textbf{ripening stage}. Seed yield is linearly related to photosynthetically active radiation that is intercepted during this phase, which aligns with our attribution results.
        Moreover, while several factors during flowering can limit the yield, rapeseed has the potential for growth after flowering, which compensates for losses of buds, flowers and pods \cite{diepenbrock2000yield}. 
        
        In the wheat dataset, significant importance is observed during the \textbf{tillering stage}, which is recognized to have great agronomic importance in cereals, as recognized by Acevedo et al. \cite{acevedo2006growth}. While it is the most important process governing canopy formation, it is also known that not all tillers produce spikes, which are the grain-bearing tip located at the top of cereal stems, as noted by Gallagher and Biscoe \cite{gallagher1978physiological}. 
        As \textbf{shoot development} progresses, spikes grow, the pseudo-stem becomes more erect, and leaf sheaths elongate. This results in a more distinguishable canopy density at satellite resolution, aligning with the high attribution of the \textbf{shoot development stage}, as illustrated in the bottom right plot of Figure \ref{fig:temp_imp_bars_GS}.
        Additionally, florets are initiated during this stage; however, it's noteworthy that less than half of these florets will complete anthesis \cite{acevedo2006growth}. As the crop progresses to the \textbf{heading stage}, the florets become ready for pollination and fertilization. It is at the \textbf{flowering stage} that the model leverages the blossoms for its predictions, as the number of flowers successfully pollinated directly influences the number of kernels per head. This observation accounts for the pronounced importance attributed to the \textbf{flowering stage}.
        During subsequent stages, the crop undergoes grain filling and moisture reduction during ripening, achieving maximum dry weight, and physiological maturity. During this period, the green color progressively fades, a change that can be captured by satellite data.

        The analysis conducted above validates our temporal importance results aggregated over growth stages, strongly supporting that our models learned agronomically meaningful cues for accurate yield predictions.

\section{Conclusion}\label{sec:conclusion}

    In this study, we used four large yield prediction datasets to explain the task of pixel-wise yield prediction.
    We demonstrated that augmenting the input data with more satellite instances or additional modalities can enhance the performance of the models. Conversely, substituting satellite bands with vegetation indices led to a degradation in the model's accuracy in most cases. Consequently, we decided against utilizing these indices to enhance the spectral interpretability of the input data.
    
    In terms of interpreting model predictions, we compared several methods and identified the \gls{svs} technique as the most robust technique.
    We proceeded with the explanation of various experiments through spectral and temporal analysis, and we showed how temporal aggregation can enhance the interpretation of attribution results and facilitates comparisons across different fields and datasets. 
    Leveraging the growth stage information enabled us to aggregate the temporal attributions into agronomically-meaningful periods, and validate our results against crop-specific domain knowledge.
    Notably, our results underscore the significance of the juvenile growth and pod filling periods across the various crops examined.

    We conducted further analysis to explain the in-field variability of the predicted yield maps, showing how the standard deviation of the attribution maps can identify the features most conducive to subfield performance.
    Additionally, we examined the correlation between attribution results and field-level performance scores. We found that monthly sampling yielded significant results, whereas using the raw time series for training failed to produce distinguishable attribution results between well- and poorly-performing fields. Under both training settings, however, aggregating over growth stages has proved to be .
    We further raised the question about the distinct attribution patterns in soybean data between Argentina and Uruguay and conducted additional data analysis to explain these differences.

\section{Future Work}
    While our study has provided valuable insights into the spectral and temporal importance of satellite data to predict crop yield, several directions remain open for investigation to further explain this task.
    We are currently conducting an in-depth exploration of attributions when utilizing intermediate fusion models, and the corresponding findings will be published separately.
    The data fusion approach is particularly relevant as it handles the static and dynamic features separately before merging them at an intermediate or late stage. 
    Moreover, other types of satellite can enhance the yield prediction task, namely the \gls{sar} data. In our future work we will explore the potential of integrating Sentinel-1 data into our modalities, facilitated by the intermediate fusion approach.
    Additionally, since multispectral data can be affected by cloud coverage, further analysis is required to assess the nature and magnitude of this impact.
    Lastly, there is a growing trend in utilizing uncertainty estimation to support the explanation of model predictions and errors. Conducting additional experiments can help us further explore this direction.

\section*{Acknowledgments}
    The research results presented are part of a large collaborative project on agricultural yield predictions, which was partly funded through the ESA InCubed Programme (\url{https://incubed.esa.int/}) as part of the project AI4EO Solution Factory (\url{https://www.ai4eo-solution-factory.de/}).
    H.Najjar acknowledges support through a scholarship from the University of Kaiserslautern-Landau.
\appendices

    \newpage
    \section{Modeling Data}\label{app:data}

        \begin{table*}[!]
\caption{Characteristics of all input features from satellite data and additional modalities used for crop yield prediction.}
\centering

\begin{tabular}{lcccclcc}
\hline
\textbf{Dynamic features} & \textbf{Source} & \textbf{Spatial Res.} & \textbf{Temporal Res.} & \text{\qquad } & \textbf{Static features} & \textbf{Source} & \textbf{Spatial Res.} \\
\hline
B01 - Coastal Aerosol & Sentinel 2 & 60 m & 5 days & & Elevation (dem) & SRTM & 30 m \\
B02 - Blue & Sentinel 2 & 10 m & 5 days & & Slope & SRTM & 30 m \\
B03 - Green & Sentinel 2 & 10 m & 5 days & & Curvature & SRTM & 30 m \\
B04 - Red & Sentinel 2 & 10 m & 5 days & & Topographic wetness index (twi) & SRTM & 30 m \\
B05 - Red Edge 1 & Sentinel 2 & 20 m & 5 days & & Aspect & SRTM & 30 m \\
B06 - Red Edge 2 & Sentinel 2 & 20 m & 5 days & & Cation exchange capacity (cec) & SoilGrids & 250 m \\
B07 - Red Edge 3 & Sentinel 2 & 20 m & 5 days & & Vol. frac. of course fragments (cfvo) & SoilGrids & 250 m \\
B08 - NIR & Sentinel 2 & 10 m & 5 days & & Nitrogen & SoilGrids & 250 m \\
B8A - Narrow NIR & Sentinel 2 & 20 m & 5 days & & Soil pH (phh2o) & SoilGrids & 250 m \\
B09 - Water vapour & Sentinel 2 & 60 m & 5 days & & Sand & SoilGrids & 250 m \\
B11 - SWIR 1 & Sentinel 2 & 20 m & 5 days & & Silt & SoilGrids & 250 m \\
B12 - SWIR 2 & Sentinel 2 & 20 m & 5 days & & Soil organic carbon (soc) & SoilGrids & 250 m \\
Max temperature & ERA5 & 30 km & Daily & & Clay & SoilGrids & 250 m \\
Mean temperature & ERA5 & 30 km & Daily & &  &  &  \\
Min temperature & ERA5 & 30 km & Daily & &  &  &  \\
Total precipitation & ERA5 & 30 km & Daily & &  &  &   \\
\hline
\end{tabular}

\label{table:features}
\end{table*}
        
        \paragraph{Yield Data} Yield maps were collected by combine harvesters, over three countries and three crop types. The data was acquired between 2016 and 2022, resulting in more than one thousand (field, season) combinations. Note that in Table \ref{table:datasets} the same field can be counted more than once in case its yield data was collected in more than one year. The data cleaning was conducted through the reprojection of all maps into a common coordinate reference system, and data points were filtered through a statistical threshold \cite{sanchezInfluenceData2023}. Maps with large inconsistencies in the recorded yield values or missing time stamp were removed. The resulting point vector data per field is rasterized into 10m resolution yield maps.
        
        \paragraph{Sentinel-2 Data} The entire spectral signal of Sentinel-2 Level-2A data with all scenes from seeding to harvesting is collected within the crop growing period. The additional \gls{scl} layer is used to identify and omit clouded pixels. Spectral bands with lower resolutions are upsampled to 10m resolution, resulting in twelve spectral bands.
        
        \paragraph{\glsentryfull{ads}} Within the spatial boundaries of each field, we collect weather data derived from the ECMWF Reanalysis (ERA5) \cite{hersbach2020era5}, soil data from SoilGrids in 250m resolution, and \gls{dem} data from NASA’s \gls{srtm} \cite{farr2000shuttle} in 30m resolution.
            Weather data is aggregated for each day at field level for minimum, maximum, and mean temperature and total precipitations.
            We use eight soil properties, i.e. cation exchange capacity (cec), volumetric fraction of course fragments (cfvo), nitrogen, soil pH (phh2o), sand, silt, soil organic carbon (soc), and clay, at depth of 0-5, 5-15, and 15-30 cm.
            For DEM, in addition to the elevation values, we derived the aspect, curvature, slope and the \gls{twi}.
            Soil and DEM data were transformed into raster images and upsampled to  a 10m resolution, using a cubic spline interpolation.

    \section{Modeling}\label{app:model}
        For all models, two \gls{lstm} layers are stacked on top of each other, with 128 hidden units each, and a dropout of 0.3\% to prevent overfitting. The output of the last time step of the second \gls{lstm} layer is fed into a sequence of operations: The first linear layer has 128 input features and 128 output features. A batch normalization is applied to the output of the first linear layer, which helps in stabilizing and accelerating the training process, followed by a ReLU activation function. The second linear layer with 128 input features and a single output feature is then applied to make the final yield prediction.

        The models are optimized using the \glsentryfull{mae} as a loss function. Additionally, it is employed for model evaluation, alongside the \glsentryfull{rmse} and the\glsentryfull{r2}. The respective formulas for each metric are as follows:
        $$
\text{MAE}=\frac{1}{n} \sum_{i=1}^n\left|y_i-\hat{y}_i\right|
$$

$$
\text{RMSE}=\sqrt{\frac{\sum_{i=1}^n\left(y_i-\hat{y}_i\right)^2}{n}}
$$

$$
\text{R}^2=1-\frac{\sum_{i=1}^n\left(y_i-\hat{y}_i\right)^2}{\sum_{i=1}^n\left(y_i-\bar{y}\right)^2} \text{, \quad } \bar{y}=\frac{1}{n} \sum_{i=1}^n y_i
$$

        Furthermore, we examine the evaluation at each cross validation fold and summarize the results in Table \ref{table:best_cv_selection}, with the selected folds highlighted in green.

\begin{table*}[]
\footnotesize
\centering 
\caption{Cross validation R² scores in each training setting, on both subfield level (SFL) and field levels (FL). Rows highlighted in green indicate the fold used for the explanation analysis of each dataset.}
\begin{tabular}{|c|cccccc|cccccc|}
 \hline
 & \multicolumn{6}{c|}{ARG-S} & \multicolumn{6}{c|}{URG-S} \\ \cline{2-13} 
 & \multicolumn{2}{c}{S2 monthly} & \multicolumn{2}{c}{S2 raw time series} & \multicolumn{2}{c|}{S2+ADS} & \multicolumn{2}{c}{S2 monthly} & \multicolumn{2}{c}{S2 raw time series} & \multicolumn{2}{c|}{S2+ADS} \\ \cline{2-13} 
\multirow{-3}{*}{Fold} & R2-SFL & R2-FL & R2-SFL & R2-FL & R2-SFL & R2-FL & R2-SFL & R2-FL & R2-SFL & R2-FL & R2-SFL & R2-FL \\ \hline

\cellcolor[HTML]{83E28E}1 & \cellcolor[HTML]{83E28E}0.68 & \cellcolor[HTML]{83E28E}0.79 & \cellcolor[HTML]{83E28E}0.70 & \cellcolor[HTML]{83E28E}0.81 & \cellcolor[HTML]{83E28E}0.69 & \cellcolor[HTML]{83E28E}0.81 & 0.22 & 0.36 & 0.34 & 0.71 & 0.25 & 0.49 \\
2 & 0.64 & 0.84 & 0.66 & 0.83 & 0.65 & 0.78 & 0.39 & 0.82 & 0.39 & 0.81 & 0.39 & 0.76 \\
3 & 0.69 & 0.81 & 0.65 & 0.78 & 0.72 & 0.88 & 0.43 & 0.74 & 0.42 & 0.70 & 0.42 & 0.66 \\
\cellcolor[HTML]{C1F0C8}4 & 0.65 & 0.85 & 0.65 & 0.80 & 0.64 & 0.77 & \cellcolor[HTML]{C1F0C8}0.48 & \cellcolor[HTML]{C1F0C8}0.78 & \cellcolor[HTML]{C1F0C8}0.51 & \cellcolor[HTML]{C1F0C8}0.89 & \cellcolor[HTML]{C1F0C8}0.49 & \cellcolor[HTML]{C1F0C8}0.88 \\
5 & 0.37 & 0.61 & 0.56 & 0.80 & 0.56 & 0.79 & 0.28 & 0.70 & 0.30 & 0.72 & 0.30 & 0.80 \\
6 & 0.53 & 0.60 & 0.58 & 0.53 & 0.55 & 0.50 & 0.40 & 0.82 & 0.40 & 0.83 & 0.40 & 0.85 \\
7 & 0.61 & 0.89 & 0.59 & 0.80 & 0.62 & 0.86 & 0.30 & 0.64 & 0.36 & 0.77 & 0.31 & 0.70 \\
8 & 0.43 & 0.71 & 0.49 & 0.66 & 0.54 & 0.83 & 0.34 & 0.60 & 0.39 & 0.65 & 0.39 & 0.66 \\
9 & 0.55 & 0.57 & 0.58 & 0.67 & 0.61 & 0.76 & 0.37 & 0.74 & 0.39 & 0.78 & 0.39 & 0.72 \\
10 & 0.60 & 0.64 & 0.67 & 0.74 & 0.69 & 0.74 & 0.47 & 0.65 & 0.49 & 0.64 & 0.49 & 0.75  \\ \hline \end{tabular}

\vspace{0.15in}

\begin{tabular}{|c|cccccc|cccccc|}
 \hline 
 & \multicolumn{6}{c|}{GER-W} & \multicolumn{6}{c|}{GER-R} \\ \cline{2-13} 
 & \multicolumn{2}{c}{S2 monthly} & \multicolumn{2}{c}{S2 raw time series} & \multicolumn{2}{c|}{S2+ADS} & \multicolumn{2}{c}{S2 monthly} & \multicolumn{2}{c}{S2 raw time series} & \multicolumn{2}{c|}{S2+ADS} \\ \cline{2-13} 
\multirow{-3}{*}{Fold} & R2-SFL & R2-FL & R2-SFL & R2-FL & R2-SFL & R2-FL & R2-SFL & R2-FL & R2-SFL & R2-FL & R2-SFL & R2-FL \\ \hline
1 & 0.33 & 0.53 & 0.39 & 0.30 & 0.45 & 0.70 & 0.52 & 0.68 & 0.48 & 0.67 & 0.57 & 0.74 \\
2 & 0.39 & 0.50 & 0.36 & 0.48 & 0.53 & 0.77 & 0.22 & 0.72 & 0.33 & 0.75 & 0.29 & 0.63 \\
\cellcolor[HTML]{C1F0C8}3 & \cellcolor[HTML]{C1F0C8}0.36 & \cellcolor[HTML]{C1F0C8}0.79 & \cellcolor[HTML]{C1F0C8}0.42 & \cellcolor[HTML]{C1F0C8}0.72 & \cellcolor[HTML]{C1F0C8}0.43 & \cellcolor[HTML]{C1F0C8}0.75 & 0.32 & 0.59 & 0.24 & 0.67 & 0.43 & 0.79 \\
4 & 0.30 & 0.68 & 0.37 & 0.75 & 0.29 & 0.74 & 0.16 & 0.25 & 0.26 & 0.42 & 0.44 & 0.51 \\
5 & 0.29 & 0.72 & 0.29 & 0.65 & 0.30 & 0.75 & 0.19 & 0.42 & 0.31 & 0.54 & 0.48 & 0.70 \\
\cellcolor[HTML]{83E28E}6 & 0.16 & 0.57 & 0.28 & 0.82 & 0.13 & 0.43 & \cellcolor[HTML]{83E28E}0.47 & \cellcolor[HTML]{83E28E}0.82 & \cellcolor[HTML]{83E28E}0.56 & \cellcolor[HTML]{83E28E}0.88 & \cellcolor[HTML]{83E28E}0.57 & \cellcolor[HTML]{83E28E}0.91 \\
7 & 0.13 & 0.31 & 0.22 & 0.06 & 0.38 & 0.55 & 0.51 & 0.81 & 0.42 & 0.51 & 0.58 & 0.73 \\
8 & 0.30 & 0.46 & 0.24 & 0.45 & 0.27 & 0.40 & 0.39 & 0.55 & 0.34 & 0.39 & 0.38 & 0.55 \\
9 & 0.25 & 0.51 & 0.24 & 0.34 & 0.06 & -0.59 & 0.29 & 0.77 & 0.39 & 0.74 & 0.39 & 0.90 \\
10 & 0.18 & -0.11 & 0.40 & 0.46 & 0.49 & 0.62 & 0.32 & 0.70 & 0.40 & 0.77 & 0.42 & 0.71 \\ \hline
\end{tabular}

\label{table:best_cv_selection}
\end{table*}

    \section{Vegetation Indices}\label{app:VIs}
        For the experiment exchanging the satellite bands with \glspl{vi}, we selected 10 indices commonly used for crop monitoring, as summarized in Table \ref{tab:VIs}

\begin{table*}[t]
\centering
\footnotesize
\caption{
    Vegetation indices used in the yield modeling and their corresponding equations. For simplification purposes, certain satellite bands are replaced as follows: B for B02, G for B03, R for B04, RE1 for B05, RE2 for B06, and N for B08.
}
\begin{tabular}{ccc}
\hline
{\color[HTML]{000000} \textbf{Vegetation Index}} & {\color[HTML]{000000} \textbf{Formula}} & {\color[HTML]{000000} \textbf{Reference}} \\ \hline
{\color[HTML]{000000} Chlorophyll   Index Green (CIG)} & {\color[HTML]{000000} (N / G) - 1.0} & {\color[HTML]{000000} \cite{CIG_CIRE_gitelson2003relationships}} \\ \hline
{\color[HTML]{000000} Chlorophyll   Index Red Edge (CIRE)} & {\color[HTML]{000000} (N / RE1) - 1} & {\color[HTML]{000000} \cite{CIG_CIRE_gitelson2003relationships}} \\ \hline
{\color[HTML]{000000} \begin{tabular}[c]{@{}c@{}}Green   Normalized Difference Vegetation Index (GNDVI)\end{tabular}} & {\color[HTML]{000000} (N - G)/(N + G)} & {\color[HTML]{000000} \cite{GNDVI_gitelson1996use}} \\ \hline
{\color[HTML]{000000} \begin{tabular}[c]{@{}c@{}}Normalized   Difference Vegetation  Index (NDVI)\end{tabular}} & {\color[HTML]{000000} (N - R)/(N + R)} & {\color[HTML]{000000} \cite{NDVI_rouse1974monitoring}} \\ \hline
{\color[HTML]{000000} \begin{tabular}[c]{@{}c@{}}Normalized   Difference Vegetation Index (NDYI)\end{tabular}} & {\color[HTML]{000000} (G - B) / (G + B)} & {\color[HTML]{000000} \cite{NDYI_sulik2016spectral}} \\ \hline
{\color[HTML]{000000} Ratio   Vegetation Index (RVI)} & {\color[HTML]{000000} RE2 / R} & {\color[HTML]{000000} \cite{RVI_birth1968measuring}} \\ \hline
{\color[HTML]{000000} \begin{tabular}[c]{@{}c@{}}Wide   Dynamic Range Vegetation  Index (WDRVI)\end{tabular}} & {\color[HTML]{000000} (0.1 * N - R) / (0.1 * N + R)} & {\color[HTML]{000000} \cite{WDRVI_gitelson2004wide}} \\ \hline
{\color[HTML]{000000} \begin{tabular}[c]{@{}c@{}}Normalized   Green Red Difference Index (NGRDI)\end{tabular}} & {\color[HTML]{000000} (G - R) / (G + R)} & {\color[HTML]{000000} \cite{NGRDI_tucker1979red}} \\ \hline
{\color[HTML]{000000} \begin{tabular}[c]{@{}c@{}}Modified Chlorophyll Absorption Ratio  Index / Optimized \\ Soil-Adjusted Vegetation Index (MCARI/OSAVI)\end{tabular}} & {\color[HTML]{000000} \begin{tabular}[c]{@{}c@{}}(((RE2 - RE1) - 0.2 * (RE2 - G)) * (RE2   / RE1)) /\\  (1.16 * (RE2 - RE1) / (RE2 + RE1 + 0.16))\end{tabular}} & {\color[HTML]{000000} \cite{MCARIO_TCARIO_wu2008estimating}} \\ \hline
{\color[HTML]{000000} \begin{tabular}[c]{@{}c@{}}Transformed Chlorophyll Absorption Ratio Index / Optimized \\ Soil-Adjusted Vegetation Index (TCARI/OSAVI)\end{tabular}} & {\color[HTML]{000000} \begin{tabular}[c]{@{}c@{}}(3 * ((RE2 - RE1) - 0.2 * (RE2 - G) *   (RE2 / RE1))) / \\ (1.16 * (RE2 - RE1) / (RE2 + RE1 + 0.16))\end{tabular}} & {\color[HTML]{000000} \cite{MCARIO_TCARIO_wu2008estimating}} \\ \hline
\end{tabular}
\label{tab:VIs}

\end{table*}

    \section{Evaluation results using the padded baseline }\label{app:padded_baseline}
\begin{figure}[htbp]
    \centering
    \includegraphics[width=\columnwidth]{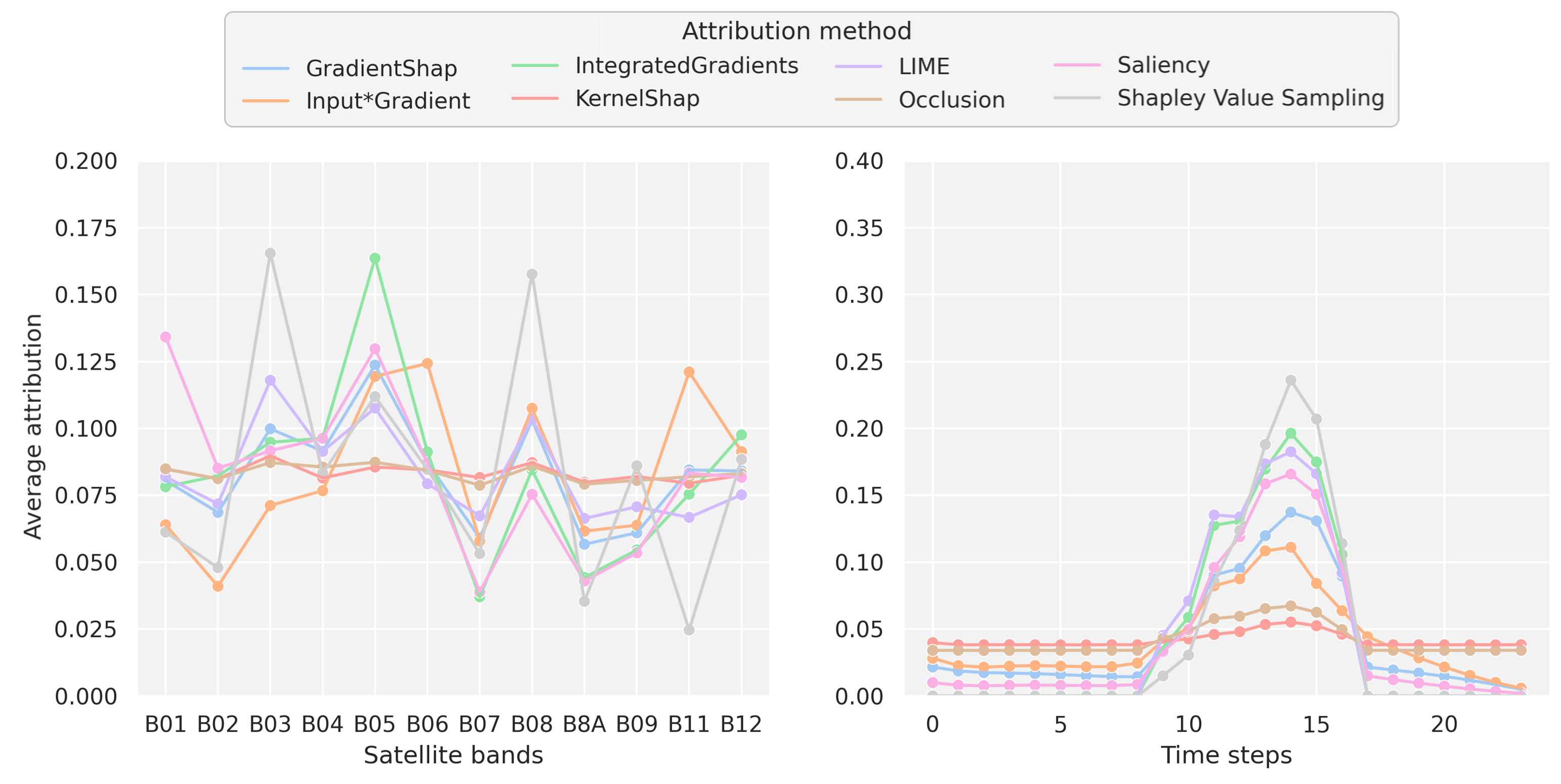}
    \caption{XAI Methods results using the padded baseline and ARG-S dataset. On the left is the spectral importance and on the right the temporal importance.}
    \label{fig:padded_baseline_spectral_temporal_importance}
\end{figure}

\begin{figure}[t]
    \centering
    \includegraphics[width=0.9\columnwidth]{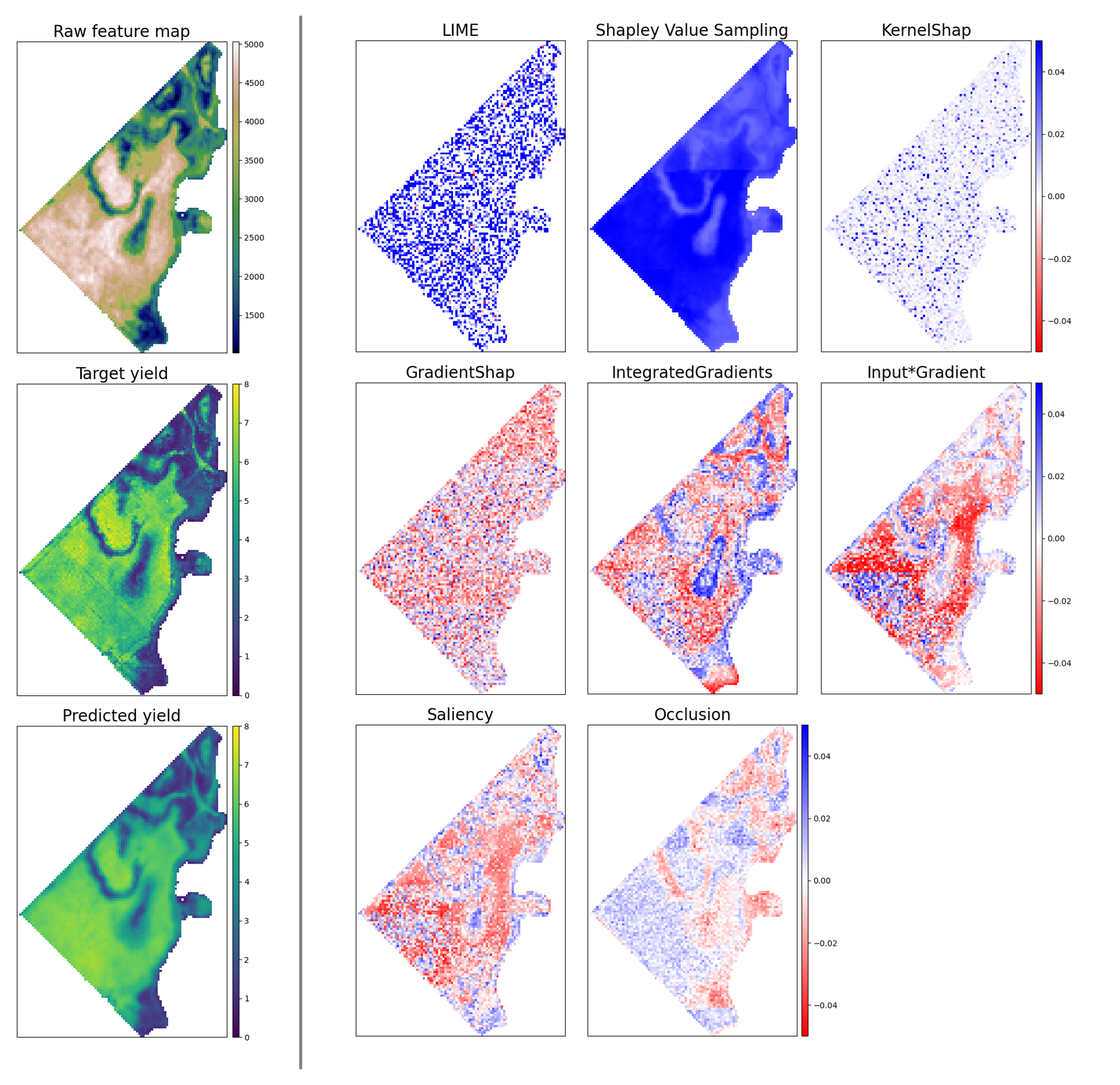}
    \caption{
        Qualitative evaluation of the feature attribution methods using the padded baseline on ARG-S dataset: Attribution maps of band B08 of time step 14 (March of the second year).
        }
    \label{fig:padded_baseline_attr_maps}
\end{figure}

\begin{figure}[t]
    \centering
    \includegraphics[width=0.7\columnwidth]{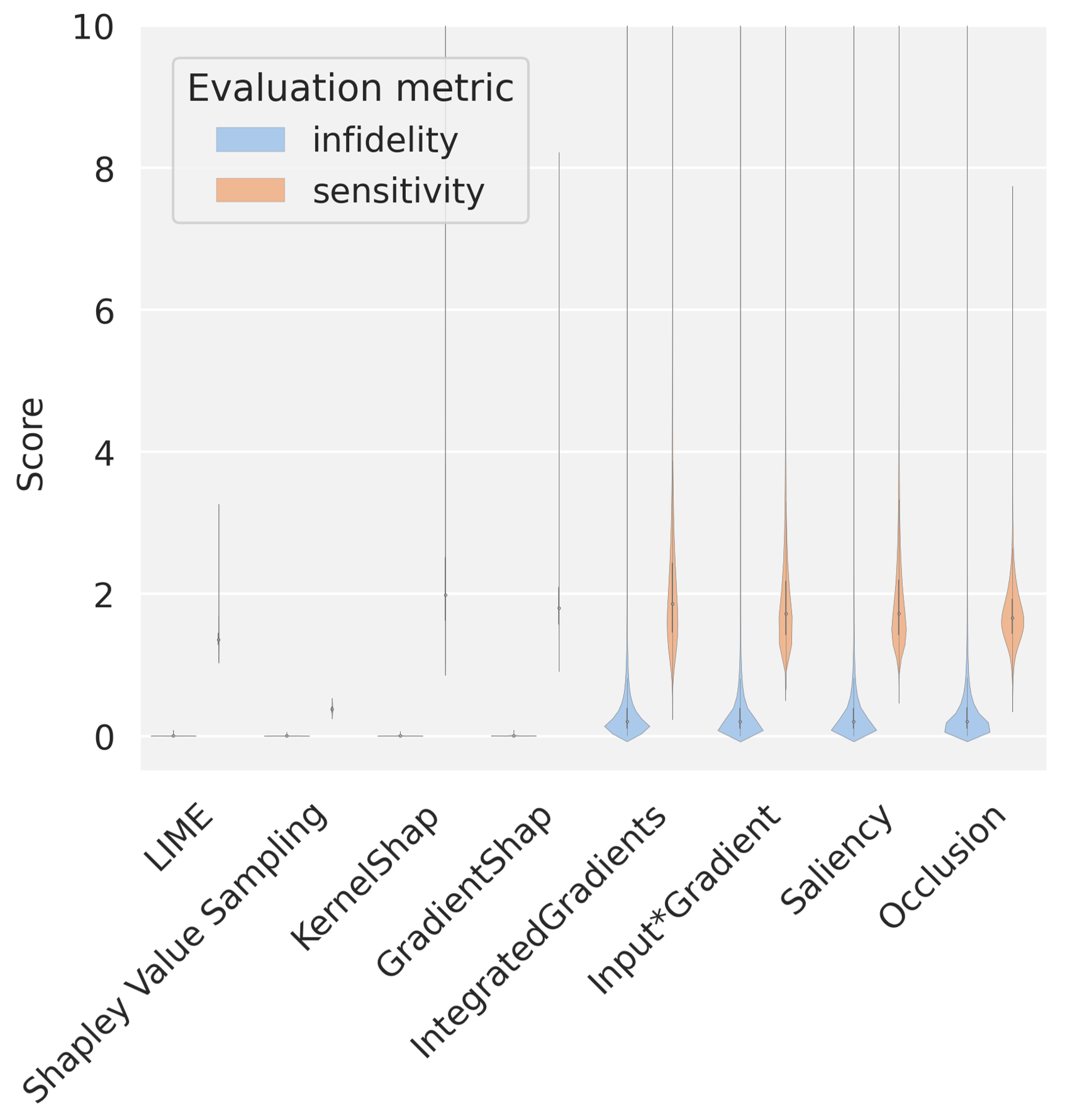}
    \caption{
        Quantitative evaluation of the feature attribution methods using the padded baseline on ARG-S dataset: Infidelity and sensitivity scores.
        }
    \label{fig:padded_baseline_eval_scores}
\end{figure}

        We report the results of the evaluation of different \gls{xai} methods using the padded baseline in Figures \ref{fig:padded_baseline_spectral_temporal_importance} and \ref{fig:padded_baseline_attr_maps} for qualitative evaluation, and in Figure \ref{fig:padded_baseline_eval_scores} for qualitative evaluation.
    
    \section{Growth Stages}\label{app:growth_stages}
    
        \begin{table}[t]
    \centering
    \footnotesize
    \caption{Growth stages adapted for each crop.}
    \begin{tabular}{lll}
        \hline
        \multicolumn{1}{c}{Rapeseed \cite{lancashire1991uniform}} & \multicolumn{1}{c}{Wheat \cite{lancashire1991uniform}} & \multicolumn{1}{c}{Soybean \cite{mcwilliams1999soybean}} \\ \hline
        Bud development & Bud development & Emergence \\
        Leaf development & Leaf development & Unifoliolate \\
        Shoot development & Tillering & 1st-25th Trifoliolate \\
        Heading & Shoot development & Beginning Bloom \\ 
        Flowering & Bolting & Full Bloom \\
        Development of fruit & Heading & Beginning Pod \\
        Ripening & Flowering & Full Pod \\
        Senescence & Development of fruit & Beginning Seed \\
         & Ripening & Full Seed \\
         & Senescence & Beginning Maturity \\
         &  & Full Maturity \\ \hline
    \end{tabular}
    \label{table:gs}
\end{table}
        
        To aggregate the temporal attributions by growth stages, we use the crop-specific scales described in Table \ref{table:gs}.
        
        Phenology data was provided by xarvio \footnote{ \href{www.xarvio.com}{www.xarvio.com}, accessed 20 March 2024.}, using in-house developed and commercially deployed growth stage models that estimate cultivar-specific growth stages of different crops on a daily base. These models are trained (as a \gls{ml} model) per country and crop, based on local cultivar-specific growth stage observations acquired in field trials (among other observations) as well as additional data sources such as weather data, to best account for local growth conditions.

    \section{Dimensionality Reduction}\label{app:dim_reduction_tsne}
\begin{figure*}[t]
    \centering  
    \includegraphics[width=0.99\textwidth]{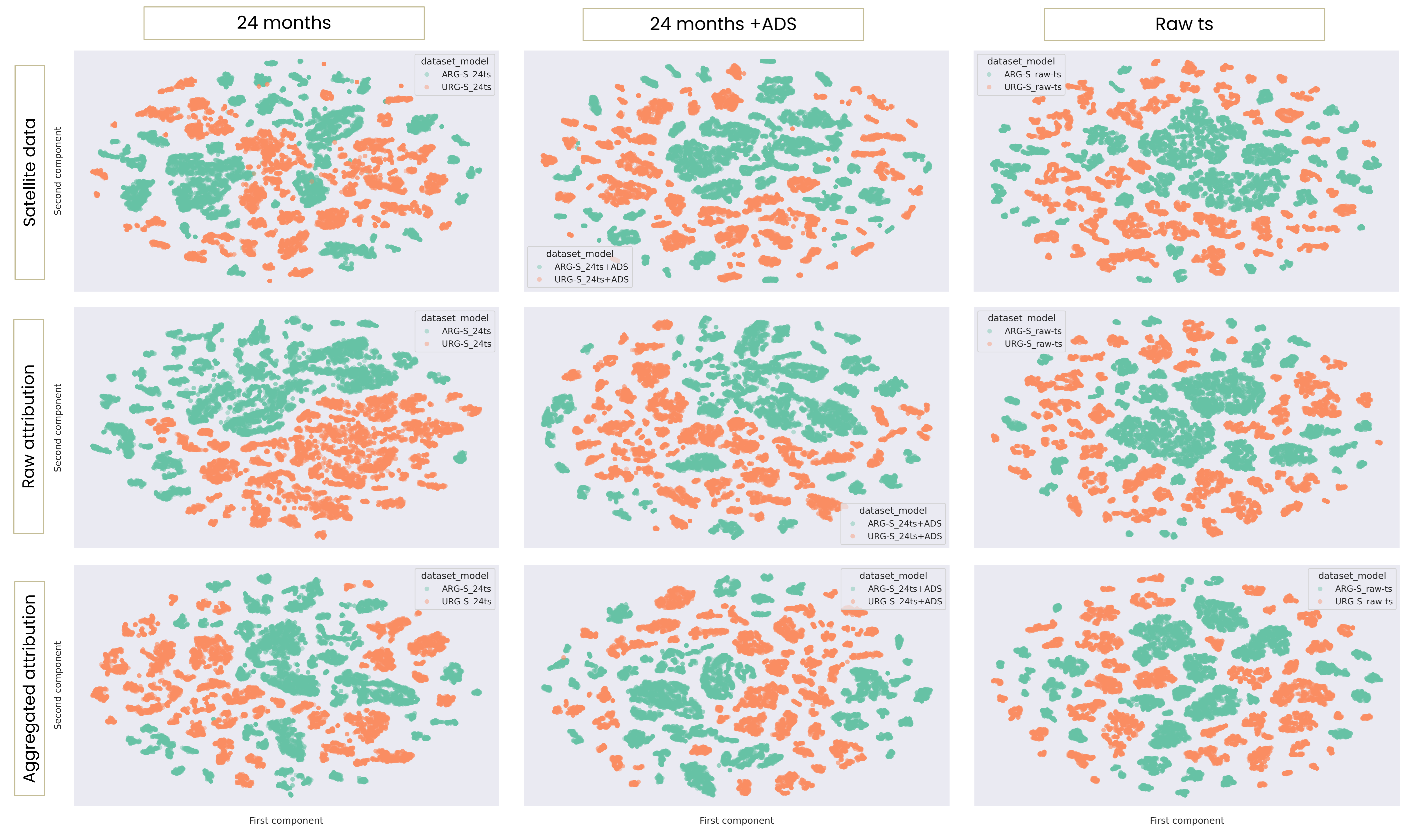}
    \caption{T-SNE results comparing Argentina and Uruguay data. The columns correspond to the different training settings, and the rows correspond to the type of vectors on which the dimensionality reduction is applied.}
    \label{fig:dim_reduction_tsne}
\end{figure*}

        Soybean data analysis through dimensionality reduction using \gls{tsne} are illustrated in Figure \ref{fig:dim_reduction_tsne}. 
        
    \newpage
    
    \quad 
    
    \newpage

\bibliographystyle{IEEEtran}
\bibliography{refs/related_work, refs/igarss23_contributions, refs/xai_methods, refs/miscellaneous, refs/vegetation_indices, refs/growth_stages}

\begin{thebibliography}{10}
\providecommand{\url}[1]{#1}
\csname url@samestyle\endcsname
\providecommand{\newblock}{\relax}
\providecommand{\bibinfo}[2]{#2}
\providecommand{\BIBentrySTDinterwordspacing}{\spaceskip=0pt\relax}
\providecommand{\BIBentryALTinterwordstretchfactor}{4}
\providecommand{\BIBentryALTinterwordspacing}{\spaceskip=\fontdimen2\font plus
\BIBentryALTinterwordstretchfactor\fontdimen3\font minus
  \fontdimen4\font\relax}
\providecommand{\BIBforeignlanguage}[2]{{%
\expandafter\ifx\csname l@#1\endcsname\relax
\typeout{** WARNING: IEEEtran.bst: No hyphenation pattern has been}%
\typeout{** loaded for the language `#1'. Using the pattern for}%
\typeout{** the default language instead.}%
\else
\language=\csname l@#1\endcsname
\fi
#2}}
\providecommand{\BIBdecl}{\relax}
\BIBdecl

\bibitem{el2022food}
M.~A. El~Mokhtar, M.~Anli, R.~B. Laouane, A.~Boutasknit, H.~Boutaj, A.~Draoui,
  L.~Zarik, and A.~Fakhech, ``Food security and climate change,'' in
  \emph{Research Anthology on Environmental and Societal Impacts of Climate
  Change}.\hskip 1em plus 0.5em minus 0.4em\relax IGI Global, 2022, pp. 44--63.

\bibitem{kemmerling2022logics}
B.~Kemmerling, C.~Schetter, and L.~Wirkus, ``The logics of war and food (in)
  security,'' \emph{Global Food Security}, vol.~33, p. 100634, 2022.

\bibitem{bhanumathi2019crop}
S.~Bhanumathi, M.~Vineeth, and N.~Rohit, ``Crop yield prediction and efficient
  use of fertilizers,'' in \emph{2019 International Conference on Communication
  and Signal Processing (ICCSP)}.\hskip 1em plus 0.5em minus 0.4em\relax IEEE,
  2019, pp. 0769--0773.

\bibitem{bondre2019prediction}
D.~A. Bondre and S.~Mahagaonkar, ``Prediction of crop yield and fertilizer
  recommendation using machine learning algorithms,'' \emph{International
  Journal of Engineering Applied Sciences and Technology}, vol.~4, no.~5, pp.
  371--376, 2019.

\bibitem{ali2022crop}
A.~M. Ali, M.~Abouelghar, A.~Belal, N.~Saleh, M.~Yones, A.~I. Selim, M.~E.
  Amin, A.~Elwesemy, D.~E. Kucher, S.~Maginan \emph{et~al.}, ``Crop yield
  prediction using multi sensors remote sensing,'' \emph{The Egyptian Journal
  of Remote Sensing and Space Science}, vol.~25, no.~3, pp. 711--716, 2022.

\bibitem{xie2022combining}
Y.~Xie, ``{Combining CERES-Wheat model, Sentinel-2 data, and deep learning
  method for winter wheat yield estimation},'' \emph{International Journal of
  Remote Sensing}, vol.~43, no.~2, pp. 630--648, 2022.

\bibitem{jeong2022predicting}
S.~Jeong, J.~Ko, and J.-M. Yeom, ``{Predicting rice yield at pixel scale
  through synthetic use of crop and deep learning models with satellite data in
  South and North Korea},'' \emph{Science of The Total Environment}, vol. 802,
  p. 149726, 2022.

\bibitem{you2017deep}
J.~You, X.~Li, M.~Low, D.~Lobell, and S.~Ermon, ``Deep gaussian process for
  crop yield prediction based on remote sensing data,'' in \emph{Proceedings of
  the AAAI conference on artificial intelligence}, vol.~31, no.~1, 2017.

\bibitem{sakamoto2014near}
T.~Sakamoto, A.~A. Gitelson, and T.~J. Arkebauer, ``{Near real-time prediction
  of US corn yields based on time-series MODIS data},'' \emph{Remote Sensing of
  Environment}, vol. 147, pp. 219--231, 2014.

\bibitem{jin2020deep}
X.~Jin, Z.~Li, H.~Feng, Z.~Ren, and S.~Li, ``{Deep neural network algorithm for
  estimating maize biomass based on simulated Sentinel 2A vegetation indices
  and leaf area index},'' \emph{The Crop Journal}, vol.~8, no.~1, pp. 87--97,
  2020.

\bibitem{leukelMachineLearning2023}
J.~Leukel, T.~Zimpel, and C.~Stumpe, ``{Machine Learning Technology for Early
  Prediction of Grain Yield at the Field Scale: {{A}} Systematic Review},''
  \emph{Computers and Electronics in Agriculture}, vol. 207, p. 107721, 2023.

\bibitem{murugananthamSystematicLiterature2022}
P.~Muruganantham, S.~Wibowo, S.~Grandhi, N.~H. Samrat, and N.~Islam, ``A
  {{Systematic Literature Review}} on {{Crop Yield Prediction}} with {{Deep
  Learning}} and {{Remote Sensing}},'' \emph{Remote Sensing}, vol.~14, no.~9,
  p. 1990, 2022.

\bibitem{neuhaus2023spurious}
Y.~Neuhaus, M.~Augustin, V.~Boreiko, and M.~Hein, ``Spurious features
  everywhere-large-scale detection of harmful spurious features in imagenet,''
  in \emph{Proceedings of the IEEE/CVF International Conference on Computer
  Vision}, 2023, pp. 20\,235--20\,246.

\bibitem{lime_ribeiro2016should}
M.~T. Ribeiro, S.~Singh, and C.~Guestrin, ``{Why should I trust you?"
  Explaining the predictions of any classifier},'' in \emph{Proceedings of the
  22nd ACM SIGKDD international conference on knowledge discovery and data
  mining}, 2016, pp. 1135--1144.

\bibitem{ig_sundararajan2017axiomatic}
M.~Sundararajan, A.~Taly, and Q.~Yan, ``Axiomatic attribution for deep
  networks,'' in \emph{International conference on machine learning}.\hskip 1em
  plus 0.5em minus 0.4em\relax PMLR, 2017, pp. 3319--3328.

\bibitem{martinez-ferrerCropYield2021}
L.~Mart{\'\i}nez-Ferrer, M.~Piles, and G.~Camps-Valls, ``Crop {{Yield
  Estimation}} and {{Interpretability With Gaussian Processes}},'' \emph{IEEE
  Geoscience and Remote Sensing Letters}, vol.~18, no.~12, pp. 2043--2047,
  2020.

\bibitem{farmonovCombiningPlanetScope2023}
N.~Farmonov, K.~Amankulova, J.~Szatm{\'a}ri, J.~Urinov, Z.~Narmanov,
  J.~Nosirov, and L.~Mucsi, ``Combining {{PlanetScope}} and {{Sentinel-2}}
  {Images with Environmental Data for Improved Wheat Yield Estimation},''
  \emph{International Journal of Digital Earth}, vol.~16, no.~1, pp. 847--867,
  2023.

\bibitem{bromsCombinedAnalysis2023}
C.~Broms, M.~Nilsson, A.~Oxenstierna, A.~Sopasakis, and K.~Åström,
  ``{Combined Analysis of Satellite and Ground Data for Winter Wheat Yield
  Forecasting},'' \emph{Smart Agricultural Technology}, vol.~3, p. 100107,
  2023.

\bibitem{shapley1953value}
L.~SHAPLEY, ``A value for n-person games,'' \emph{Contributions to the Theory
  of Games}, pp. 307--317, 1953.

\bibitem{svs_strumbelj2010efficient}
E.~Strumbelj and I.~Kononenko, ``An efficient explanation of individual
  classifications using game theory,'' \emph{The Journal of Machine Learning
  Research}, vol.~11, pp. 1--18, 2010.

\bibitem{celik2023explainable}
M.~F. Celik, M.~S. Isik, G.~Taskin, E.~Erten, and G.~Camps-Valls,
  ``{Explainable Artificial Intelligence for Cotton Yield Prediction with
  Multisource Data},'' \emph{IEEE Geoscience and Remote Sensing Letters}, 2023.

\bibitem{nori2019interpretml}
H.~Nori, S.~Jenkins, P.~Koch, and R.~Caruana, ``{Interpretml: A unified
  framework for machine learning interpretability},'' \emph{arXiv preprint
  arXiv:1909.09223}, 2019.

\bibitem{wolaninEstimatingUnderstanding2020}
A.~Wolanin, G.~Mateo-Garc{\'\i}a, G.~Camps-Valls, L.~G{\'o}mez-Chova,
  M.~Meroni, G.~Duveiller, Y.~Liangzhi, and L.~Guanter, ``{Estimating and
  Understanding Crop Yields with Explainable Deep Learning} in the {{Indian
  Wheat Belt}},'' \emph{Environmental research letters}, vol.~15, no.~2, p.
  024019, 2020.

\bibitem{huberExtremeGradient2022}
F.~Huber, A.~Yushchenko, B.~Stratmann, and V.~Steinhage, ``{Extreme Gradient
  Boosting for yield estimation compared with Deep Learning approaches},''
  \emph{Computers and Electronics in Agriculture}, vol. 202, p. 107346, 2022.

\bibitem{paudel2023interpretability}
D.~Paudel, A.~de~Wit, H.~Boogaard, D.~Marcos, S.~Osinga, and I.~N.
  Athanasiadis, ``Interpretability of deep learning models for crop yield
  forecasting,'' \emph{Computers and Electronics in Agriculture}, vol. 206, p.
  107663, 2023.

\bibitem{de201925}
A.~De~Wit, H.~Boogaard, D.~Fumagalli, S.~Janssen, R.~Knapen, D.~van Kraalingen,
  I.~Supit, R.~van~der Wijngaart, and K.~van Diepen, ``{25 years of the WOFOST
  cropping systems model},'' \emph{Agricultural systems}, vol. 168, pp.
  154--167, 2019.

\bibitem{FAPAR}
\BIBentryALTinterwordspacing
{Copernicus GLS,2020}. {Fraction of Absorbed Photosynthetically Active
  Radiation.} [Online]. Available:
  \url{https://land.copernicus.eu/global/products/fapar}
\BIBentrySTDinterwordspacing

\bibitem{isik2023interpretable}
M.~S. Isik, M.~F. Celik, and E.~Erten, ``Interpretable cotton yield prediction
  model using earth observation time series,'' in \emph{IGARSS 2023-2023 IEEE
  International Geoscience and Remote Sensing Symposium}.\hskip 1em plus 0.5em
  minus 0.4em\relax IEEE, 2023, pp. 3442--3445.

\bibitem{mateo-sanchisInterpretableLongShort2023}
A.~Mateo-Sanchis, J.~E. Adsuara, M.~Piles, J.~Munoz-Mar{\'\i}, A.~Perez-Suay,
  and G.~Camps-Valls, ``Interpretable {{Long-Short Term Memory Networks}} for
  {{Crop Yield Estimation}},'' \emph{IEEE Geoscience and Remote Sensing
  Letters}, vol.~20, pp. 1--5, 2023.

\bibitem{russwurm2020self}
M.~Ru{\ss}wurm and M.~K{\"o}rner, ``Self-attention for raw optical satellite
  time series classification,'' \emph{ISPRS journal of photogrammetry and
  remote sensing}, vol. 169, pp. 421--435, 2020.

\bibitem{hochreiter1997long}
S.~Hochreiter and J.~Schmidhuber, ``Long short-term memory,'' \emph{Neural
  computation}, vol.~9, no.~8, pp. 1735--1780, 1997.

\bibitem{fischer2021global}
G.~Fischer, F.~Nachtergaele, H.~Van~Velthuizen, F.~Chiozza, G.~Franceschini,
  M.~Henry, D.~Muchoney, and S.~Tramberend, \emph{Global agro-ecological zones
  v4-model documentation}.\hskip 1em plus 0.5em minus 0.4em\relax Food \&
  Agriculture Org., 2021.

\bibitem{sanchezInfluenceData2023}
C.~Sanchez, D.~Pathak, M.~Miranda, P.~Helber, B.~Bischke, P.~Habelitz,
  H.~Najjar, F.~Mena, J.~Siddamsetty, D.~Arenas, M.~Vollmer, M.~Charfuelan,
  M.~Nuske, and A.~Dengel, ``Influence of {{Data Cleaning Techniques}} on
  {{Sub-Field Yield Predictions}},'' in \emph{{{IGARSS}} 2023 - 2023 {{IEEE
  International Geoscience}} and {{Remote Sensing Symposium}}}, 2023.

\bibitem{pathakPredictingCrop2023}
D.~Pathak, M.~Miranda, F.~Mena, C.~Sanchez, P.~Helber, B.~Bischke, P.~Habelitz,
  H.~Najjar, J.~Siddamsetty, D.~Arenas, M.~Vollmer, M.~Charfuelan, M.~Nuske,
  and A.~Dengel, ``Predicting {{Crop Yield With Machine Learning}}: {{An
  Extensive Analysis Of Input Modalities And Models On}} a {{Field}} and
  {{Subfield Level}},'' in \emph{{{IGARSS}} 2023 - 2023 {{IEEE International
  Geoscience}} and {{Remote Sensing Symposium}}}, 2023.

\bibitem{najjarFeatureAttribution2023}
H.~Najjar, P.~Helber, B.~Bischke, P.~Habelitz, C.~Sanchez, F.~Mena, M.~Miranda,
  D.~Pathak, J.~Siddamsetty, D.~Arenas, M.~Vollmer, M.~Charfuelan, M.~Nuske,
  and A.~Dengel, ``Feature {{Attribution Methods For Multivariate Time-Series
  Explainability In Remote Sensing}},'' in \emph{{{IGARSS}} 2023 - 2023 {{IEEE
  International Geoscience}} and {{Remote Sensing Symposium}}}, 2023.

\bibitem{saliency_simonyan2014deep}
K.~Simonyan, A.~Vedaldi, and A.~Zisserman, ``{Deep inside convolutional
  networks: visualising image classification models and saliency maps},'' in
  \emph{Proceedings of the International Conference on Learning Representations
  (ICLR)}.\hskip 1em plus 0.5em minus 0.4em\relax ICLR, 2014.

\bibitem{shrikumar2016not}
A.~Shrikumar, P.~Greenside, A.~Shcherbina, and A.~Kundaje, ``{Not just a black
  box: Learning important features through propagating activation
  differences},'' \emph{arXiv preprint arXiv:1605.01713}, 2016.

\bibitem{shap_lundberg2017unified}
S.~M. Lundberg and S.-I. Lee, ``A unified approach to interpreting model
  predictions,'' \emph{Advances in neural information processing systems},
  vol.~30, 2017.

\bibitem{occ_zeiler2014visualizing}
M.~D. Zeiler and R.~Fergus, ``Visualizing and understanding convolutional
  networks,'' in \emph{Computer Vision--ECCV 2014}.\hskip 1em plus 0.5em minus
  0.4em\relax Springer, 2014, pp. 818--833.

\bibitem{yeh2019fidelity}
C.-K. Yeh, C.-Y. Hsieh, A.~Suggala, D.~I. Inouye, and P.~K. Ravikumar, ``On the
  (in) fidelity and sensitivity of explanations,'' \emph{Advances in Neural
  Information Processing Systems}, vol.~32, 2019.

\bibitem{guidotti2018survey}
R.~Guidotti, A.~Monreale, S.~Ruggieri, F.~Turini, F.~Giannotti, and
  D.~Pedreschi, ``A survey of methods for explaining black box models,''
  \emph{ACM computing surveys (CSUR)}, vol.~51, no.~5, pp. 1--42, 2018.

\bibitem{lancashire1991uniform}
P.~D. Lancashire, H.~Bleiholder, T.~V.~D. Boom, P.~Langel{\"u}ddeke, R.~Stauss,
  E.~Weber, and A.~Witzenberger, ``A uniform decimal code for growth stages of
  crops and weeds,'' \emph{Annals of applied Biology}, vol. 119, no.~3, pp.
  561--601, 1991.

\bibitem{mcwilliams1999soybean}
D.~A. McWilliams, D.~R. Berglund, and G.~Endres, \emph{Soybean growth and
  management quick guide}, 1999.

\bibitem{sudmanns2020assessing}
M.~Sudmanns, D.~Tiede, H.~Augustin, and S.~Lang, ``{Assessing global Sentinel-2
  coverage dynamics and data availability for operational Earth observation
  (EO) applications using the EO-Compass},'' \emph{International journal of
  digital earth}, vol.~13, no.~7, pp. 768--784, 2020.

\bibitem{mena2023common}
F.~Mena, D.~Arenas, M.~Nuske, and A.~Dengel, ``{Common Practices and Taxonomy
  in Deep Multi-view Fusion for Remote Sensing Applications},'' \emph{arXiv
  preprint arXiv:2301.01200}, 2023.

\bibitem{alvarez2018towards}
D.~Alvarez~Melis and T.~Jaakkola, ``Towards robust interpretability with
  self-explaining neural networks,'' \emph{Advances in neural information
  processing systems}, vol.~31, 2018.

\bibitem{bhatt2020evaluating}
U.~Bhatt, A.~Weller, and J.~M. Moura, ``Evaluating and aggregating
  feature-based model explanations,'' \emph{arXiv preprint arXiv:2005.00631},
  2020.

\bibitem{lundberg2018consistent}
S.~M. Lundberg, G.~G. Erion, and S.-I. Lee, ``Consistent individualized feature
  attribution for tree ensembles,'' \emph{arXiv preprint arXiv:1802.03888},
  2018.

\bibitem{dang2021autumn}
C.~Dang, Y.~Liu, H.~Yue, J.~Qian, and R.~Zhu, ``{Autumn crop yield prediction
  using data-driven approaches:-support vector machines, random forest, and
  deep neural network methods},'' \emph{Canadian journal of remote sensing},
  vol.~47, no.~2, pp. 162--181, 2021.

\bibitem{nevavuori2020crop}
P.~Nevavuori, N.~Narra, P.~Linna, and T.~Lipping, ``{Crop yield prediction
  using multitemporal UAV data and spatio-temporal deep learning models},''
  \emph{Remote sensing}, vol.~12, no.~23, p. 4000, 2020.

\bibitem{ma2021corn}
Y.~Ma, Z.~Zhang, Y.~Kang, and M.~{\"O}zdo{\u{g}}an, ``Corn yield prediction and
  uncertainty analysis based on remotely sensed variables using a bayesian
  neural network approach,'' \emph{Remote Sensing of Environment}, vol. 259, p.
  112408, 2021.

\bibitem{jiang2020deep}
H.~Jiang, H.~Hu, R.~Zhong, J.~Xu, J.~Xu, J.~Huang, S.~Wang, Y.~Ying, and
  T.~Lin, ``{A deep learning approach to conflating heterogeneous geospatial
  data for corn yield estimation: A case study of the US Corn Belt at the
  county level},'' \emph{Global change biology}, vol.~26, no.~3, pp.
  1754--1766, 2020.

\bibitem{thomson2002elevation}
A.~M. Thomson, R.~A. Brown, S.~J. Ghan, R.~C. Izaurralde, N.~J. Rosenberg, and
  L.~R. Leung, ``{Elevation dependence of winter wheat production in eastern
  Washington State with climate change: A methodological study},''
  \emph{Climatic Change}, vol.~54, pp. 141--164, 2002.

\bibitem{kumudini2010soybean}
S.~Kumudini, ``Soybean growth and development.'' in \emph{The soybean: botany,
  production and uses}.\hskip 1em plus 0.5em minus 0.4em\relax CABI Wallingford
  UK, 2010, pp. 48--73.

\bibitem{gay1980physiological}
S.~Gay, D.~Egli, and D.~Reicosky, ``Physiological aspects of yield improvement
  in soybeans 1,'' \emph{Agronomy Journal}, vol.~72, no.~2, pp. 387--391, 1980.

\bibitem{smith1986selection}
J.~R. Smith and R.~L. Nelson, ``Selection for seed-filling period in soybean
  1,'' \emph{Crop science}, vol.~26, no.~3, pp. 466--469, 1986.

\bibitem{diepenbrock2000yield}
W.~A. Diepenbrock, ``Yield analysis of winter oilseed rape (brassica napus l.):
  a review,'' \emph{Field crops research}, vol.~67, no.~1, pp. 35--49, 2000.

\bibitem{acevedo2006growth}
E.~Acevedo, P.~Silva, and H.~Silva, ``Growth and wheat physiology,
  development,'' \emph{Laboratory of Soil-Plant-Water Relations. Faculty of
  Agronomy and Forestry Sciences. University of Chile. Casilla}, vol. 1004,
  2006.

\bibitem{gallagher1978physiological}
J.~Gallagher and P.~Biscoe, ``A physiological analysis of cereal yield. ii.
  partitioning of dry matter,'' \emph{Agricultural progress}, vol.~53, pp.
  51--70, 1978.

\bibitem{hersbach2020era5}
H.~Hersbach, B.~Bell, P.~Berrisford, S.~Hirahara, A.~Hor{\'a}nyi,
  J.~Mu{\~n}oz-Sabater, J.~Nicolas, C.~Peubey, R.~Radu, D.~Schepers
  \emph{et~al.}, ``{The ERA5 global reanalysis},'' \emph{Quarterly Journal of
  the Royal Meteorological Society}, vol. 146, no. 730, pp. 1999--2049, 2020.

\bibitem{farr2000shuttle}
T.~G. Farr and M.~Kobrick, ``{Shuttle Radar Topography Mission produces a
  wealth of data},'' \emph{Eos, Transactions American Geophysical Union},
  vol.~81, no.~48, pp. 583--585, 2000.

\bibitem{CIG_CIRE_gitelson2003relationships}
A.~A. Gitelson, Y.~Gritz, and M.~N. Merzlyak, ``Relationships between leaf
  chlorophyll content and spectral reflectance and algorithms for
  non-destructive chlorophyll assessment in higher plant leaves,''
  \emph{Journal of plant physiology}, vol. 160, no.~3, pp. 271--282, 2003.

\bibitem{GNDVI_gitelson1996use}
A.~A. Gitelson, Y.~J. Kaufman, and M.~N. Merzlyak, ``{Use of a green channel in
  remote sensing of global vegetation from EOS-MODIS},'' \emph{Remote sensing
  of Environment}, vol.~58, no.~3, pp. 289--298, 1996.

\bibitem{NDVI_rouse1974monitoring}
J.~W. Rouse, R.~H. Haas, J.~A. Schell, D.~W. Deering \emph{et~al.},
  ``{Monitoring vegetation systems in the Great Plains with ERTS},'' \emph{NASA
  Spec. Publ}, vol. 351, no.~1, p. 309, 1974.

\bibitem{NDYI_sulik2016spectral}
J.~J. Sulik and D.~S. Long, ``Spectral considerations for modeling yield of
  canola,'' \emph{Remote Sensing of Environment}, vol. 184, pp. 161--174, 2016.

\bibitem{RVI_birth1968measuring}
G.~S. Birth and G.~R. McVey, ``Measuring the color of growing turf with a
  reflectance spectrophotometer 1,'' \emph{Agronomy Journal}, vol.~60, no.~6,
  pp. 640--643, 1968.

\bibitem{WDRVI_gitelson2004wide}
A.~A. Gitelson, ``Wide dynamic range vegetation index for remote quantification
  of biophysical characteristics of vegetation,'' \emph{Journal of plant
  physiology}, vol. 161, no.~2, pp. 165--173, 2004.

\bibitem{NGRDI_tucker1979red}
C.~J. Tucker, ``Red and photographic infrared linear combinations for
  monitoring vegetation,'' \emph{Remote sensing of Environment}, vol.~8, no.~2,
  pp. 127--150, 1979.

\bibitem{MCARIO_TCARIO_wu2008estimating}
C.~Wu, Z.~Niu, Q.~Tang, and W.~Huang, ``{Estimating chlorophyll content from
  hyperspectral vegetation indices: Modeling and validation},''
  \emph{Agricultural and forest meteorology}, vol. 148, no. 8-9, pp.
  1230--1241, 2008.

\end{thebibliography}

\newpage

\begin{IEEEbiographynophoto}{Hiba Najjar}  
received her M.Sc. degree in Applied Mathematics from the Ecole Nationale Supérieure des Mines de Nancy, Nancy, France, in 2021. She is pursuing her Ph.D. degree at the University
of Kaiserslautern-Landau, Kaiserslautern, Germany. Her research interests include explainable AI, remote sensing, and agricultural applications.
\end{IEEEbiographynophoto}

\begin{IEEEbiographynophoto}{Miro Miranda}
received a Bachelor's degree in Agricultural Sciences from the University of Giessen and a Master’s degree in Life Science Informatics from the University of Bonn in 2021. After a research stay at the University of Bonn, he joined the German Research Center for Artificial Intelligence, pursuing his Ph.D. degree in computer science at the University of Kaiserslautern-Landau. 
\end{IEEEbiographynophoto}

\begin{IEEEbiographynophoto}{Marlon Nuske} received his Master's and PhD degree in Physics from University of Hamburg in 2015 and 2020, respectively. He is now working at DFKI in Kaiserslautern as a Senior Researcher leading the Earth and Space Applications team since 2021. His research interests lie in machine learning applications in Earth Observation, data fusion, hybrid modelling techniques and physics-aware machine learning.
\end{IEEEbiographynophoto}

\begin{IEEEbiographynophoto}{Ribana Roscher}
(Member,~IEEE) received the Dipl.Ing. and Ph.D. degrees in geodesy from the University of Bonn, Bonn, Germany, in 2008 and
2012, respectively. Until 2022, she was a Junior Professor of Remote Sensing with the University of Bonn. She was a Post-Doctoral Researcher with the University of Bonn, the Julius Kuehn Institute, Siebeldingen, Germany, Freie Universität Berlin, Berlin, Germany,
and Humboldt Innovation, Berlin. In 2015, she was a Visiting Researcher with the Fields Institute, Toronto, ON, Canada. Since 2022, she has been a Professor of Data Science for Crop Systems with the University of Bonn and leads the same-titled group at the Institute of Bio- and Geosciences (IBG)-2, Forschungszentrum Jülich, Jülich, Germany.
\end{IEEEbiographynophoto}

\begin{IEEEbiographynophoto}{Andreas Dengel}
is the Executive Director of DFKI in Kaiserslautern. He received Diploma from University of Kaiserslautern in 1986 and a PhD from the University of Stuttgart in 1989. He worked for IBM, Siemens and Xerox Parc and became a Professor at the Department of Computer Science at TU Kaiserslautern in 1993. Since 2009, he has held another professorship (kyakuin) in the Department of Computer Science and Intelligent Systems at Osaka Metropolitan University with teaching and examination rights. At this university, he was also named "Distinguished Honorary Professor" (tokubetu eiyo kyoju) in March 2018, a distinction received by only five researchers within 135 years. Andreas has chaired numerous international conferences and serves on the editorial boards of international journals and book series. He has written or edited 14 books and is the (co-)author of more than 600 peer-reviewed scientific publications, many of which have received Best Paper Awards. To date, he has supervised more than 500 doctoral, master's, and bachelor's theses. In addition to a number of keynote lectures at international meetings and conferences, Andreas has given invited technical presentations at numerous prestigious universities and research institutions. These include MIT, Stanford University, PARC, UC Berkeley, CMU, London School of Economics, Cambridge University, ATR, NII, Tokyo University, Chinese Academy of Science, and Google Research or MS Research. Also, activities as a lecturer, e.g., in the Joint Executive MBA Program at Johannes Gutenberg University in Mainz, the University of Texas in Austin and Dongbei University of Finance and Economics in Dalian, China, are part of Andreas' previous activities. He is an IAPR fellow and a member of the member of the National Academy of Science and Technology (acatech). In addition to the honors already mentioned, Andreas has received other important recognitions for his work, some of which are mentioned below. Back in 1997, Andreas already received one of the most prestigious personal science awards in Germany, the Alcatel/SEL Award for Technical Communication, for his scientific achievements. In 2019, he also received the Outstanding Achievement Award from the Int'l Conference on Document Analysis and Recognition (ICDAR) in Sydney, Australia. Finally, in 2021, Andreas was awarded the oldest Japanese order, the "The Order of the Rising Sun, Gold Rays with Neck Ribbon" in the name of His Majesty Emperor Naruhito and in 2022 he was awarded the Order of Merit of the State of Rhineland-Palatinate.
\end{IEEEbiographynophoto}

\end{document}